\documentclass{article} 
\usepackage{iclr2020_conference,times}

\usepackage{amsmath,amsfonts,bm}

\def\eqref#1{equation~\ref{#1}}

\def\1{\bm{1}}

\DeclareMathAlphabet{\mathsfit}{\encodingdefault}{\sfdefault}{m}{sl}
\SetMathAlphabet{\mathsfit}{bold}{\encodingdefault}{\sfdefault}{bx}{n}

\usepackage{graphicx}
\usepackage{hyperref}
\usepackage{url}
\usepackage{algorithm}
\usepackage{algorithmic}

\newcommand{\norm}[1]{\left\lVert #1 \right\rVert}

\newcommand{\G}[1]{G\left( #1 ; \theta_G \right)}
\newcommand{\D}[1]{D\left( #1 ; \theta_D \right)}
\newcommand{\Dz}{\Delta z}
\newcommand{\pG}{\theta_G}

\newcommand{\pD}{\theta_D}
\newcommand{\fz}{f(z)}
\newcommand{\fzp}{f(z')}
\newcommand{\ff}{f(z;\theta_D, \theta_G)}
\newcommand{\ffe}[3]{f(z #1 ;\theta_D #2, \theta_G #3)}
\newcommand{\expt}[2]{\mathbb{E}_{#1} \left[ #2 \right]}

\newcounter{dbaCounter}

\title{LOGAN: Latent Optimisation for Generative Adversarial Networks}

\author{Yan Wu, Jeff Donahue, David Balduzzi, Karen Simonyan, Timothy Lillicrap\\
DeepMind\\
London, UK \\
\texttt{\{yanwu,jeffdonahue,dbalduzzi,simonyan,countzero\}@google.com} \\
}

\iclrfinalcopy 
\begin{document}

\maketitle

\begin{abstract}
Training generative adversarial networks requires balancing of delicate adversarial dynamics. 
Even with careful tuning, training may diverge or end up in a bad equilibrium with dropped modes. 
In this work, we improve CS-GAN with natural gradient-based latent optimisation and show that it improves adversarial dynamics by enhancing interactions between the discriminator and the generator. 
Our experiments demonstrate that latent optimisation can significantly improve GAN training, obtaining state-of-the-art performance for the ImageNet ($128 \times 128$) dataset. Our model achieves an Inception Score (IS) of $148$ and an Fr\'{e}chet Inception Distance (FID) of $3.4$, an improvement of $17\%$ and $32\%$ in IS and FID respectively, compared with the baseline BigGAN-deep model with the same architecture and number of parameters.
\end{abstract}


\section{Introduction}

Generative Adversarial Nets (GANs) are implicit generative models that can be trained to match a given data distribution.  GANs were originally developed by~\citet{goodfellow2014generative} for image data.
As the field of generative modelling has advanced, GANs remain at the frontier, generating high-fidelity images at large scale \citep{brock2018large,karras2019style}. 
However, despite growing insights into the dynamics of GAN training, much of the progress in GAN-based image generation come from network architecture improvements \citep{radford2015unsupervised,zhang2019self}, or regularisation of particular parts of the model \citep{miyato2018spectral,miyato2018cgans}.

Build on the compressed sensing view of GANs (CS-GAN; \citealp{wu2019deep}), we improve the efficacy of latent optimisation in adversarial games, using natural gradient descent to optimise the latent variable (usually denoted $z$) towards the direction favoured by the discriminator during training.
This results in a scalable and easy to implement approach that improves the dynamic interaction between the discriminator and the generator.
We generally call these approaches latent optimised GANs (LOGAN).

\begin{figure}[htb]
    \centering
    \begin{tabular}{c c}
    \includegraphics[width=0.45\linewidth]{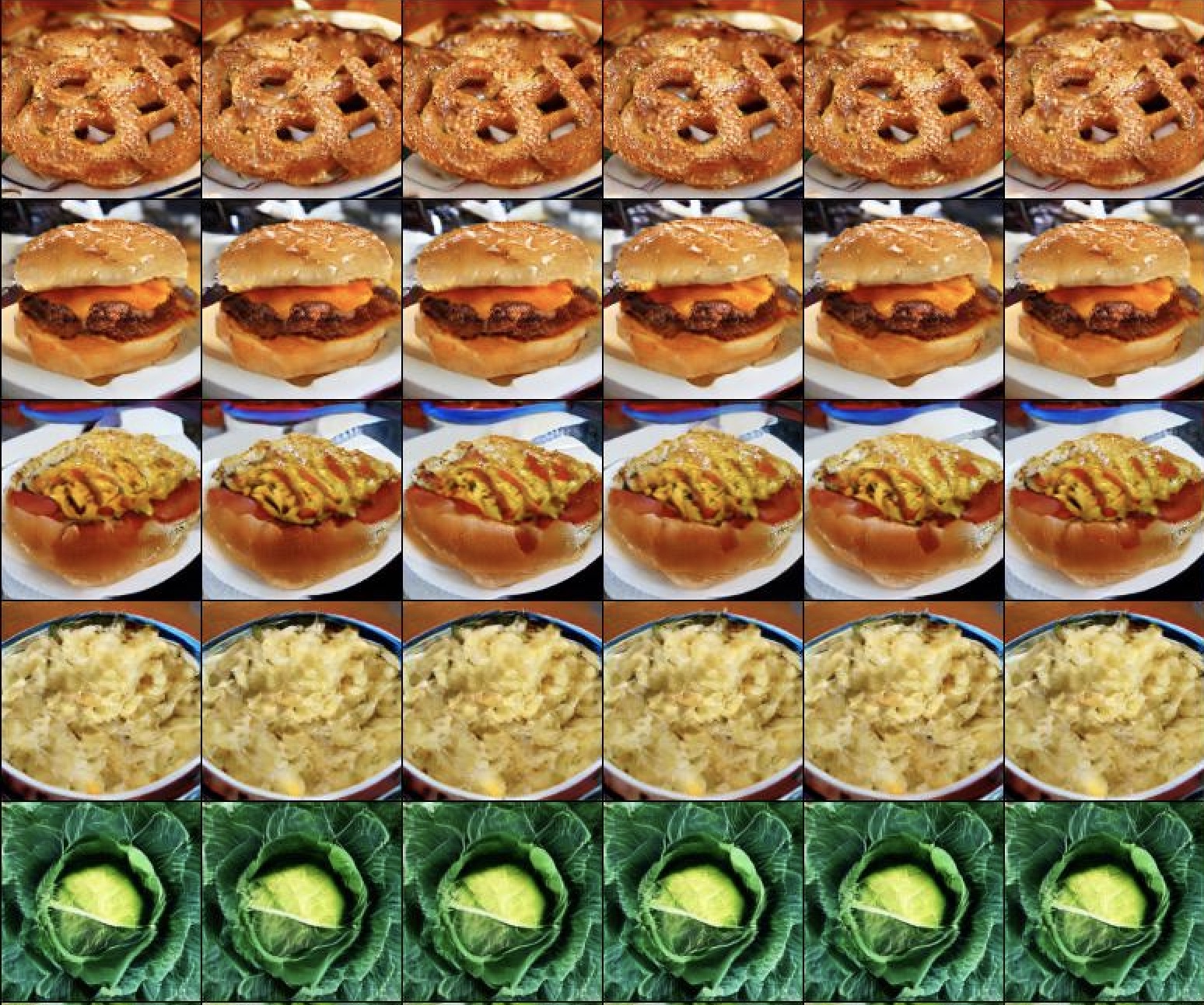} &
    \includegraphics[width=0.45\linewidth]{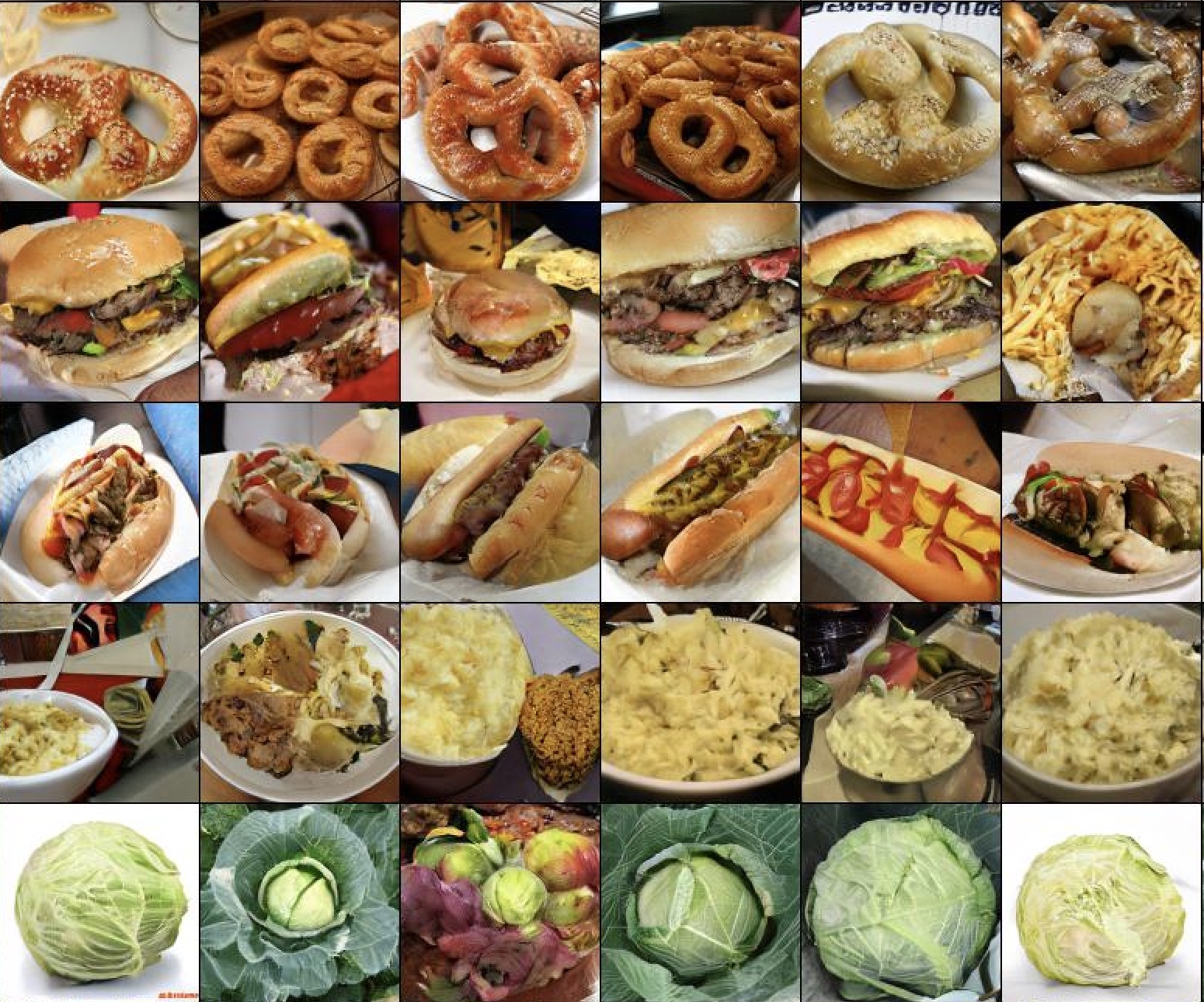} \\
    (\textbf{a}) &
    (\textbf{b}) \\
    \end{tabular}
    \caption{Samples from BigGAN-deep (\textbf{a}) and LOGAN (\textbf{b}) with similarly high IS. Samples from the two panels were drawn from truncation levels corresponding to points C and D in Figure~\ref{fig:model-truncation} \textbf{b} respectively. (FID/IS: (\textbf{a}) 27.97/259.4, (\textbf{b}) 8.19/259.9)}
    \label{fig:samples-high-IS}
\end{figure}

\begin{figure}[ht]
    \centering
    \begin{tabular}{cc}
    \includegraphics[width=0.45\linewidth]{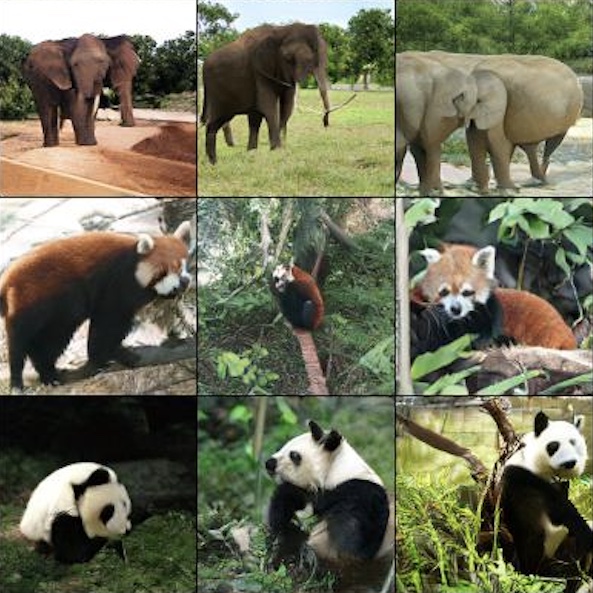} &
    \includegraphics[width=0.45\linewidth]{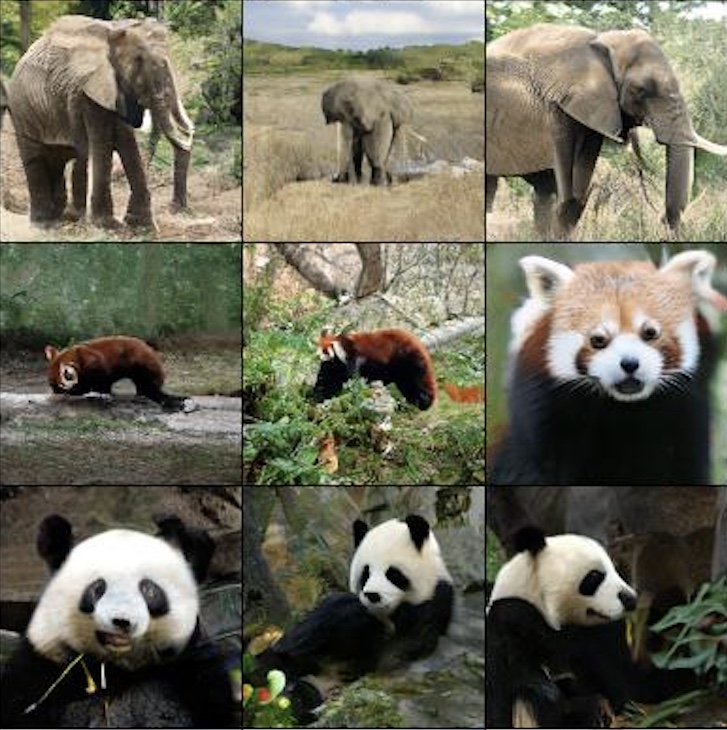} \\
    (\textbf{a}) &
    (\textbf{b}) 
    \end{tabular}
    \caption{Samples from BigGAN-deep (\textbf{a}) and LOGAN (\textbf{b}) with  similarly low FID. Samples from the two panels were drawn from truncation levels corresponding to points A and B in Figure~\ref{fig:model-truncation} \textbf{b} respectively. (FID/IS: (\textbf{a}) 5.04/126.8, (\textbf{b}) 5.09/217.0)}
    \label{fig:samples-low-FIDs}
\end{figure}

To summarise our contributions:
\begin{enumerate}
    \item We propose an improved, efficient approach to latent optimisation using natural gradient descent.
    \item Our algorithm improves the state-of-the-art BigGAN~\citep{brock2018large} by a significant margin, without introducing any architectural change, resulting in higher quality images and more diverse samples (see Table~\ref{tab:scores}, Figure~\ref{fig:samples-high-IS} and \ref{fig:samples-low-FIDs}).
    \item To provide theoretical insight, we analyse latent optimisation in GANs from the perspective of differentiable games~\citep{balduzzi2018mechanics}. We argue that latent optimisation can be viewed as improving the dynamics of adversarial training.
\end{enumerate}

\section{Background}

\subsection{Notation}
We use $\theta_D$ and $\theta_G$ to denote the vectors representing parameters of the generator and discriminator. We use $x$ for images, and $z$ for the latent source generating an image. We use prime $'$ to denote a variable after one update step, e.g., $\pD' = \pD - \alpha \frac{\partial \ff}{\partial \pD}$. $p(x)$ and $p(z)$ denote the data distribution and source distribution respectively. $\expt{p(x)}{f(x)}$ indicates taking the expectation of function $f(x)$ over the distribution $p(x)$.

\subsection{Generative Adversarial Nets}

\begin{table}[tb]
\caption{Comparison of model scores. BigGAN-deep results are reproduced from \cite{brock2018large}. ``baseline" indicates our reproduced BigGAN-deep with small modifications. The 3rd and 4th columns are from the gradient descent (GD, ablated) and natural gradient descent (NGD) versions of LOGAN respectively. We report the Inception Score (IS, higher is better, \citealt{salimans2016improved}) and Fr{\'e}chet Inception Distance (FID, lower is better, \citealt{heusel2017gans}).}
\label{sample-table}
\begin{center}
\begin{tabular}{l|ll}
         & FID & IS \\
\hline
BigGAN-deep  & $5.7 \pm 0.3$ & $124.5 \pm 2.0$ \\
baseline & $4.92 \pm 0.05$ & $126.6 \pm 1.3$ \\
LOGAN (GD) & $4.86 \pm 0.09$ & $127.7 \pm 3.5$ \\
LOGAN (NGD) & $\mathbf{3.36 \pm 0.14}$ & $\mathbf{148.2 \pm 3.1}$
\end{tabular}
\end{center}
\label{tab:scores}
\end{table}

A GAN consists of a generator that generates image $x = G(z; \theta_G)$ from a latent source $z \sim p(z)$, and a discriminator that scores the generated images as $D(x; \theta_D)$ \citep{goodfellow2014generative}. Training GANs involves an adversarial game: while the discriminator tries to distinguish generated samples $x = \G{z}$ from data $x \sim p(x)$, the generator tries to fool the discriminator.
This procedure can be summarised as the following min-max game:
\begin{equation}
    \min_{\pD} \max_{\pG} \quad \expt{x \sim p(x)}{h_D(\D{x})} + \expt{z \sim p(z)}{h_G(\D{\G{z}})}
    \label{eq:minmax}
\end{equation}
The exact form of $h(\cdot)$ depends on the choice of loss function \citep{goodfellow2014generative,arjovsky2017wasserstein,nowozin2016f}. To simplify our presentation and analysis, we use the Wasserstein loss~\citep{arjovsky2017wasserstein}, so that $h_D(t) = -t$ and $h_G(t) = t$. Our experiments with BigGAN-deep use the hinge loss \citep{lim2017geometric,tran2017hierarchical}, which is identical to this form in its linear regime. 
Our analysis can be generalised to other losses as in previous theoretical work (e.g., \citealt{arora2017generalization}).
To simplify notation, we abbreviate $\ff = \D{\G{z}}$, which may be further simplified as $\fz$ when the explicit dependency on $\pD$ and $\pG$ can be omitted.

Training GANs requires carefully balancing updates to $D$ and $G$, and is sensitive to both architecture and algorithm choices \citep{salimans2016improved,radford2015unsupervised}. A recent milestone is BigGAN (and BigGAN-deep, \citealt{brock2018large}), which pushed the boundary of high fidelity image generation by scaling up GANs to an unprecedented level. BigGANs use an architecture based on residual blocks \citep{he2016deep}, in combination with regularisation mechanisms and self-attention \citep{saxe2013exact, miyato2018spectral,zhang2019self}.

Here we aim to improve the adversarial dynamics during training. We focus on the second term in eq.~\ref{eq:minmax} which is at the heart of the min-max game. For clarity, we explicitly write the losses for $D$ as $L_D(z) = h_D(f(z))$ and $G$ as $L_G(z) = h_G(f(z))$, so the total loss vector can be written as
\begin{equation}
    L(z) = \left[ L_D(z), L_G(z) \right]^T = \left[ \fz, -\fz \right]^T
    \label{eq:mult-loss}
\end{equation}

Computing the gradients with respect to $\theta_{D}$ and $\theta_{G}$ gives the following vector field, which \emph{cannot} be expressed as the gradient of any single function \citep{balduzzi2018mechanics}:

\begin{equation}
    g = \left[ \frac{\partial L_D(z)}{\partial \theta_D},  \frac{\partial L_G(z)}{\partial \theta_G} \right]^T = \left[ \frac{\partial f(z)}{\partial \theta_D}, - \frac{\partial f(z)}{\partial \theta_G} \right]^T \label{eq:sg}
\end{equation}
The fact that $g$ is not the gradient of a function implies that gradient updates in GANs can exhibit cycling behaviour which can slow down or prevent convergence. 
\citet{balduzzi2018mechanics} refer to vector fields of this form as the \emph{simultaneous gradient}. Although many GAN models use alternating update rules (e.g., \citealt{goodfellow2014generative,brock2018large}), following the gradient with respect to $\pD$ and $\pG$ alternatively in each step, they share the same problem from gradients of this form. 
Therefore, we use the simpler simultaneous gradient (eq.~\ref{eq:sg}) for our analysis (see also \citealt{mescheder2017,mescheder2018training}).

\begin{figure}[ht]
    \centering
    \begin{tabular}{cc}
    \includegraphics[width=0.3\linewidth]{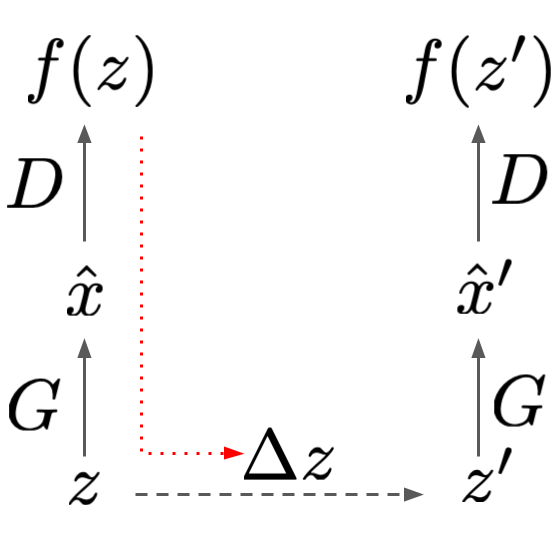}  &
    \includegraphics[width=0.5\linewidth]{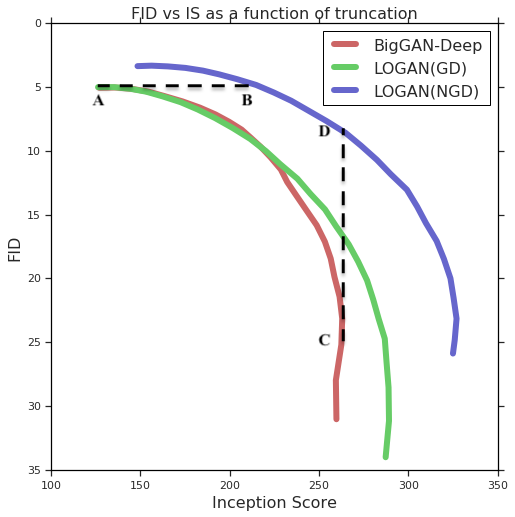} \\
    (\textbf{a}) &
    (\textbf{b}) 
    \end{tabular}
    \caption{(\textbf{a}) Schematic of LOGAN. We first compute a forward pass through $G$ and $D$ with a sampled latent $z$. Then, we use gradients from the generator loss (dashed red arrow) to compute an improved latent, $z'$. After we use this optimised latent code in a second forward pass, we compute gradients of the discriminator back through the latent optimisation into the model parameters $\theta_{D}$, $\theta_{G}$. We use these gradients to update the model.
    (\textbf{b}) Truncation curves illustrate the FID/IS trade-off for each model by altering the range of the noise source $p(z)$. GD: gradient descent. NGD: natural gradient descent. Points A, B, C, D correspond to samples shown in Figure~\ref{fig:samples-high-IS} and \ref{fig:samples-low-FIDs}.}
    \label{fig:model-truncation}
\end{figure}

\subsection{Latent Optimised GANs}

\begin{algorithm}[tb]
   \caption{Latent Optimised GANs with Automatic Differentiation}
\begin{algorithmic}
   \STATE {\bfseries Input:} data distribution $p(x)$, latent distribution $p(z)$, $\D{\cdot}$, $\G{\cdot}$, learning rate $\alpha$, batch size $N$
   \REPEAT
   \STATE Initialise discriminator and generator parameters $\pD$, $\pG$
   \FOR{$i=1$ {\bfseries to} $N$}
       \STATE Sample $z \sim p(z)$, $x \sim p(x)$
       \STATE Compute the gradient $\frac{\partial D(G(z))}{\partial z}$ and use it to obtain $\Delta z$ from eq.~\ref{eq:lo} (GD) or eq.~\ref{eq:ng-g} (NGD)
       \STATE Optimise the latent $z' \leftarrow \left[ z + \Delta z\right]$, $[\cdot]$ indicates clipping the value between $-1$ and $1$ 
       \STATE Compute generator loss $L_G^{(i)} = - D(G(z'))$
       \STATE Compute discriminator loss $L_D^{(i)} = D(G(z')) - D(x)$
   \ENDFOR
   \STATE Compute batch losses
       $L_G = \frac{1}{N} \sum_{i=1}^N L_G^{(i)}$ and
       $L_D = \frac{1}{N} \sum_{i=1}^N L_D^{(i)}$
   \STATE Update $\pD$ and $\pG$ with the gradients $\frac{\partial L_D}{\partial \pD}$, $\frac{\partial L_G}{\partial \pG}$
   \UNTIL{reaches the maximum training steps}
\end{algorithmic}
\label{alg}
\end{algorithm}

Inspired by compressed sensing \citep{candes2006stable,donoho2006compressed}, \citet{wu2019deep} introduced latent optimisation for GANs. Latent optimisation exploits knowledge from $D$ to refine the latent source $z$. Intuitively, the gradient $\nabla_z\fz = \frac{\partial \fz}{\partial z}$ points in the direction that better satisfies the discriminator $D$, which implies better samples. Therefore, instead of using the randomly sampled $z \sim p(z)$, \citet{wu2019deep} uses the optimised latent
\begin{equation}
    \Delta z = \alpha \, \frac{\partial \fz}{\partial z} \qquad z' = z + \Delta z
    \label{eq:lo}
\end{equation}
in eq.~\ref{eq:minmax} for training \footnote{Although multiple gradient descent steps can be employed for optimising $z$, we found one step works well in training and justify this choice in section \ref{sec:analysis}.}. 

Historically, compressed sensing has been developed as a signal processing technique mostly without any concern on training. However, here we emphasise the influence of this procedure on training, which we will show dominates the effects on large scale models --- in contrast, the run-time optimisation that is central in compressed sensing may be unnecessary after training. Therefore, we call this type of models latent-optimised GANs (LOGAN) to avoid any confusion, except when explicitly referring to the results from \cite{wu2019deep}. Latent optimisation has been shown to improve the stability of training as well as the final performance for medium-sized models such as DCGANs and Spectral Normalised GANs \citep{radford2015unsupervised,miyato2018spectral}. The general algorithm is summarised in Algorithm~\ref{alg} and illustrated in Figure~\ref{fig:model-truncation} \textbf{a}. However, we found that the potential of latent optimisation remained largely untapped in this setting, and develop the natural gradient descent form of latent update in Section~\ref{sec:logan-ngd}.

\section{Analysis of the Algorithm}
\label{sec:analysis}
To understand how latent optimisation interacts with GAN training, we analyse LOGAN as a differentiable game following \citet{balduzzi2018mechanics, gemp:18, letcher2019differentiable}.
The Appendix~\ref{app:analysis} provides a complementary analysis form the perspective of stochastic approximation \citep{heusel2017gans,borkar1997stochastic}. 
We can explicitly compute the gradients for the discriminator and generator at $z'$ after one step of latent optimisation by differentiating $\left[L_D(z'), L_G(z') \right]^T = \left[ f(z'), -f(z') \right]$ (where $z' = z + \Delta z$ from eq.~\ref{eq:lo}):
\begin{align}
    \left[\frac{d L_D}{d \theta_D}, \frac{d L_G}{d \theta_G} \right]^T &=  \left[\frac{\partial f(z')}{\partial \theta_D} + \left(\frac{\partial \Delta z}{\partial \theta_D} \right)^T \, \frac{\partial f(z')}{\partial \Delta z} , \, -\frac{\partial f(z')}{\partial \theta_G} - \left(\frac{\partial \Delta z}{\partial \theta_G} \right)^T \, \frac{\partial f(z')}{\partial \Delta z} \right]^T 
    \\
    &=  \left[ \frac{\partial f(z')}{\partial \theta_D} + \alpha \, \left(\frac{\partial^2 \fz}{\partial z \partial \theta_D} \right)^T \, \frac{\partial f(z')}{\partial z'} , \,
    - \frac{\partial f(z')}{\partial \theta_G} - \alpha \, \left(\frac{\partial^2 \fz}{\partial z \partial \theta_G} \right)^T \, \frac{\partial f(z')}{\partial z'}
   \right]^T
    \label{eq:lo-g}
\end{align}
In both equations, the first terms represent how $f(z')$ depends on the parameters directly, which also appear in the gradients from vanilla GANs (eq.~\ref{eq:sg}). However, the second terms are introduced from latent optimisation, accounting for how $f(z')$ depends on the parameters via the change $\Delta z$. For the second equality, we substitute $\Dz = \alpha \, \frac{\partial \fz}{\partial z}$ as the gradient-based update of $z$ and use $\frac{\partial f(z')}{\partial \Delta z} = \frac{\partial f(z')}{\partial z'}$.
Further differentiating $\Delta z$ results in the second-order terms $\left(\frac{\partial^2 \fz}{\partial z \partial \theta_D} \right)^T$ and $\left(\frac{\partial^2 \fz}{\partial z \partial \theta_G} \right)^T$.
The original GAN's gradient (eq.~\ref{eq:sg}) does not include any second-order term, since $\Delta z = 0$ without latent optimisation. LOGAN computes these extra terms by automatic differentiation when back-propagating through the latent optimisation process (see Algorithm \ref{alg}). 

\subsection{Relation with SGA}

\citet{balduzzi2018mechanics, gemp:18} proposed Symplectic Gradient Adjustment (SGA) to improve the dynamics of gradient-based methods in adversarial games. SGA addresses an important problem with gradient-based optimisation in GANs: the vector-field generated by the losses of the discriminator and generator is not a gradient vector field. It follows that gradient descent is not guaranteed to find a local optimum and can cycle, which can slow down convergence or lead to phenomena like mode collapse and mode hopping. 

For a game with gradient $g$ (eq.~\ref{eq:sg}), the Hessian is the second order derivatives with respect to the parameters, $H = \nabla_\theta g$. SGA uses the adjusted gradient
\begin{equation}
    g^* = g + \lambda \, A^T \, g 
\end{equation}
where $\lambda$ is a positive constant and $A = \frac{1}{2} (H - H^T)$ is the anti-symmetric component of the Hessian. Applying SGA to GANs yields the adjusted updates (see Appendix~\ref{app:analysis}.1 for details):
\begin{equation}
\begin{split}
    g^* = \left[ \frac{\partial \fz}{\partial \theta_D} 
    + \lambda \, \left(\frac{\partial^2 f(z)}{\partial \theta_G \, \partial \theta_D} \right)^T \, \frac{\partial f(z)}{\partial \theta_G},\,
      -\frac{\partial f(z)}{\partial \theta_G}
    + \lambda \, \left(\frac{\partial^2 f(z)}{\partial \theta_D \, \partial \theta_G} \right)^T \, \frac{\partial f(z)}{\partial \theta_D}
    \right]^T
\end{split}
    \label{eq:adjust-g}
\end{equation}
Compared with $g$ in eq.~\ref{eq:sg}, the adjusted gradient $g^*$ has second-order terms reflecting the interactions between $D$ and $G$. SGA significantly improves GAN training in simple examples \citep{balduzzi2018mechanics}, allowing faster and more robust convergence to stable fixed points (local Nash equilibria). 
Unfortunately, SGA is expensive to scale because computing the second-order derivatives with respect to all parameters is expensive. It remains unclear whether SGA can be incorporated into very large scale models using more efficient implementation (e.g., Hessian-vector products from modified back propagation \cite{pearlmutter1994fast}).

The SGA updates in eq.~\ref{eq:adjust-g} and the LOGAN updates in eq.~\ref{eq:lo-g} are strikingly similar, suggesting that the latent step used by LOGAN reduces the negative effects of cycling by introducing a symplectic gradient adjustment into the optimisation procedure. The role of the latent step can be formalised in terms of a third player, whose goal is to help the generator (see appendix~\ref{app:analysis} for details).  
Crucially, latent optimisation approximates SGA using only second-order derivatives with respect to the latent $z$ and parameters of the discriminator and generator \emph{separately}. The second order terms involving parameters of both the discriminator and the generator -- which are expensive to compute -- are not used. In short, with a simple modification of the original GAN training algorithm, latent optimisation couples the gradients of the discriminator and generator in a way similar to SGA.

\subsection{Relation with Unrolled GANs}
\label{app:unroll}

In addition, latent optimisation can be seen as unrolling GANs \citep{DBLP:journals/corr/MetzPPS16} in the space of the latent source, rather than the parameters. Unrolling in the latent space has the advantages that:
\begin{enumerate}
    \item LOGAN is more scalable than Unrolled GANs because it avoids unrolling the parameter updating process, which is prohibitively expensive for models with a large number of parameters.
    \item While unrolling the update of $D$ only affects the parameters of $G$ (as in \citealt{DBLP:journals/corr/MetzPPS16}), latent optimisation effects  both $D$ and $G$ as shown in eq.~\ref{eq:lo-g}.
\end{enumerate}
We next formally present this connection by first showing that SGA can be seen as approximating Unrolled GANs \citep{DBLP:journals/corr/MetzPPS16}. For the update $\pD' = \pD + \Delta \pD$, we have the Taylor expansion approximation at $\pD$:
\begin{equation}
\begin{split}
    \ffe{}{+ \Delta \pD}{} \approx \ff + \left( \frac{\partial \ff}{\partial \pD}\right)^T\, \Delta \pD
\end{split}
\end{equation}
Substitute the gradient descent parameter update $\Delta \pD = - \alpha \frac{\partial \ff}{\partial \pD}$, and take the derivatives with respect to $\pG$ on both sides:
\begin{equation}
\begin{split}
    \frac{\partial \ffe{}{+ \Delta \pD}{}}{\partial \pG} \approx \frac{\partial \ff}{\partial \pG} - 2\alpha \, \left( \frac{\partial^2 \ff}{\partial \pD \partial \pG}\right)^T \, \frac{\partial \ff}{\partial \pD}
\end{split}
    \label{eq:ugan-g}
\end{equation}
which has the same form as eq.~\ref{eq:adjust-g} (taking the negative sign).
Compared with the exact gradient from the unroll:
\begin{equation}
    \frac{\partial \ffe{}{+ \Delta \pD}{}}{\partial \pG} = \frac{\partial \ffe{}{'}{}}{\partial \pG} - 2\alpha \, \left( \frac{\partial^2 \ff}{\partial \pD \partial \pG}\right)^T\, \frac{\partial \ffe{}{'}{}}{\partial (\pD')}
\end{equation}
The approximation in eq.~\ref{eq:ugan-g} comes from using $\frac{\partial \ffe{}{}}{\partial \pD} \approx \frac{\partial \ffe{}{'}{}}{\partial \pD'}$ and $\frac{\partial \ff}{\partial \pG} \approx \frac{\partial \ffe{}{'}}{\partial \pG}$ as a result of additional linear approximation.

At this point, unrolling $D$ update only affects
$\pD$. Although it is expensive to unroll both $D$ and $G$, in principle, we can unroll $G$ update and compute the gradient of $\pD$ similarly using $\Delta \pG = \alpha \frac{\partial \ff}{\partial \pG}$:
\begin{equation}
    \frac{\partial \ffe{}{}{+ \Delta \pG}}{\partial \pD} \approx \frac{\partial \ff}{\partial \pD} + 2\alpha \, \left( \frac{\partial^2 \ff}{\partial \pG \partial \pD}\right)^T\, \frac{\partial \ff}{\partial \pG}
    \label{eq:ugan-d}
\end{equation}
which gives us the same update rule as SGA (eq.~\ref{eq:adjust-g}). This correspondence based on first order Taylor expansion is unsurprising, as SGA is based on linearising the adversarial dynamics \citep{balduzzi2018mechanics}.

Therefore, given the previous section, we can view LOGAN as further approximating Unrolled GAN, by unrolling the update of latent source $z$ instead of the parameters. Although the information from $z$ is limited compared with all the parameters, the intuition from Unrolled GANs applies here: unrolling the update of $z$ gives $D$ and $G$ extra information to react to their opponents, thus avoiding the circular behaviour.


\section{LOGAN with Natural Gradient Descent}
\label{sec:logan-ngd}

Our analysis explains why latent optimisation may help GAN training. In practice, we expect more benefit from latent optimisation from \emph{stronger} optimiser for $z$, which can better capture the coupling between $D$ and $G$.
\cite{wu2019deep} only used basic gradient descent (GD) with a fixed step-size. This choice limits the size $\Dz$ can take: in order not to overshoot when the curvature is large, the step size would be too conservative when the curvature is small. We hypothesis that GD is more detrimental for larger models, which have complex loss surfaces with highly varying curvatures. Consistent with this hypothesis, we observed only marginal improvement over the baseline using GD (section~\ref{sec:exp-ablation}, Table~\ref{tab:scores}, Figure~\ref{fig:model-truncation} \textbf{b}).

In this work, we propose using natural gradient descent (NGD, \citealt{amari1998natural}) for latent optimisation. NGD is an approximate second-order optimisation method, and has been applied successfully in many domains \citep{pascanu2013revisiting,martens2014new}. By using the positive semi-definite (PSD) Gauss-Newton matrix to approximate the (possibly negative definite) Hessian, NGD often works even better than exact second-order methods. NGD is expensive in high dimensional parameter spaces, even with approximations \citep{martens2014new}. However, we demonstrate that it is efficient for latent optimisation, even in very large models.

Given the gradient of $z$, $g = \frac{\partial \fz}{\partial z}$, NGD computes the update as
\begin{equation}
    \Delta z = \alpha \, F^{-1} \, g
    \label{eq:ngd}
\end{equation}
where the Fisher information matrix $F$ is defined as
\begin{equation}
    F =  \expt{p(t|z)}{\nabla \ln p(t|z) \, \nabla \ln p(t|z)^T}
\end{equation}
The log-likelihood function $\ln p(t|z)$ typically corresponds to commonly used error functions such as the cross entropy loss. This correspondence is not necessary when we interpret NGD as an approximate second-order method, as has long been done \citep{martens2014new}. Nevertheless, Appendix~\ref{app:poisson-ll} provides a Poisson log-likelihood interpretation for the hinge loss commonly used in GANs \citep{lim2017geometric,tran2017hierarchical}.
An important difference between latent optimisation and commonly seen scenarios using NGD is that the expectation over the condition ($z$) is absent. Since each $z$ is only responsible for generating one image, it only minimises the loss $L_G(z)$ for this particular instance.

More specifically, we use the \emph{empirical} Fisher $F'$ with Tikhonov damping, as in TONGA~\citep{roux2008topmoumoute}
\begin{equation}
    F' = g \cdot g^T + \beta \, I
    \label{eq:e-fisher}
\end{equation}
$F'$ is cheaper to compute compared with the full Fisher, since $g$ is already available.
The \emph{damping factor} $\beta$ regularises the step size, which is important when $F'$ only poorly approximates the Hessian or when the Hessian changes too much across the step.
Using the Sherman-Morrison formula, the NGD update can be simplified into the following closed form:
\begin{equation}
    \Delta z = \alpha \, \left( \frac{I}{\beta} - \frac{g\,g^T}{\beta^2 + \beta \, g^T\,g} \right) \, g = \frac{\alpha}{\beta + \|g\|^2} \, g
    \label{eq:ng-g}
\end{equation}
which does not involve any matrix inversion. Thus, NGD adapts the step size according to the curvature estimate $c = \frac{1}{\beta + \|g\|^2}$.
When $\beta$ is small, NGD normalises the gradient by its squared L2-norm.
NGD automatically smooths the scale of updates by down-scaling the gradients as their norm grows, which also contributes to the smoothed norms of updates (Appendix~\ref{app:analysis}.2). Since the NGD update remains proportional to $g$, our analysis based on gradient descent in section~\ref{sec:analysis} still holds. 

\subsection*{Additional Regularisation}

Various regularisation techniques are often necessary to ensure the stable training of GANs. Here we highlight two of them that we found particularly useful in combination with LOGAN. First, we found regularising the Euclidean norm of optimisation step,
\begin{equation}
    R_z = w_r \cdot \norm{\Dz}_2^2
    \label{eq:z_reg}
\end{equation}
where the scalar weight $w_r$ is a parameter, as introduced by \citet{wu2019deep} is necessary, especially for large models. This term is added to both the generator loss and discriminator loss in training. \citet{wu2019deep} suggested this term is related to optimal transport; more recently, \citet{tanaka2019discriminator} formalised this connection in Discriminator Optimal Transport (DOT). We left the exact connection between our work and DOT to future investigation, but here note that while DOT improves evaluation performance, our method mainly focuses on training. Consequently, although the regulariser $R_z$ shares the same form as that in DOT, they function differently: it regularises the update of parameters here, but the latent code $z$ in DOT.

In addition, we found it is more stable to optimise only a portion $c$ of $z$, leaving some of its elements completely random, which can be seem an additional damping mechanism while preserve more randomness from the latent source.

\section{Experiments and Analysis}
\label{sec:experiments}

We tested our algorithm for both medium (DCGAN, \citealt{radford2015unsupervised,miyato2018spectral}) and large scale (BigGAN, \citealt{brock2018large}) models. We use the standard hyper-parameter settings for each GAN model, without further optimising them with LOGAN. We performed grid-search over the four parameters introduced in LOGAN: the latent step size $\alpha$, damping factor $\beta$, the regularisation weight $w_r$, and the portion of $z$ being optimised as $c$. Details of the grid search are summarised in Appendix~\ref{app:hyper}. Additional empirical analysis of latent optimisation is presented in Appendix \ref{app:analyse-logan}.

\subsection{Experiments with DCGAN on CIFAR}
\label{app:dcgan}

To verify if our proposed NGD optimiser works well for latent optimisation, we first test LOGAN at more moderate scales for direct comparison with \citet{wu2019deep} using basic GD. Here we apply latent optimisation on Spectral Normalised GANs (SN-GANs, \citealt{miyato2018spectral}). 

The experiments follows the same basic setup and hyper-parameter settings as the CS-GAN in \cite{wu2019deep}. There is no class conditioning in this model. With NGD, we use a large step size of $\alpha = 0.9$ and the damping factor $\beta = 0.1$ for optimising $z$. We found the weight of $0.1$ for the regulariser $R_z$ (eq.~\ref{eq:z_reg}), and optimising $80\%$ of the latent source worked best for SN-GANs. All other parameters are same as in \citealt{wu2019deep}.

In addition, we found running extra latent optimisation steps benefited evaluation, so we use ten steps of latent optimisation in evaluation for results in this section, although the models were still trained with a single optimisation step. This is different from in larger models, where optimisation is unnecessary in evaluation (see section~\ref{sec:biggan-basic} for more details).

\begin{figure}[htb]
    \centering
    \begin{tabular}{cc}
    \includegraphics[width=0.45\linewidth]{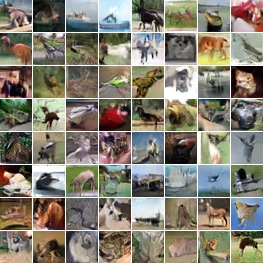} &
    \includegraphics[width=0.45\linewidth]{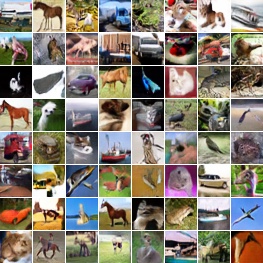} \\
    (\textbf{a}) &
     (\textbf{b})
    \end{tabular}
    \caption{(\textbf{a}) Samples from SN-GAN. (\textbf{b}) Samples from LOGAN.}
    \label{fig:app_samples}
\end{figure}

Table~\ref{tab:dcgan-scores} shows the FID and IS alongside SN-GAN and CS-CAN which used the same architecture. The scores are computed based on $10,000$ samples following the same procedure as in \citet{wu2019deep}. We observe that NGD brought significant improvement over CS-GAN (i.e., LOGAN with GD for optimising $z$). Compared with the baseline SN-GAN model without employing any latent optimisation, there is an improvement of $16.8\%$ in IS and $39.6\%$ in FID.
Figure~\ref{fig:app_samples} compares random samples from these two models. Overall, samples from LOGAN (NGD) have higher contrasts and sharper contours.

\begin{table}[htb]
\caption{Comparison of Scores. The first and second columns are reproduced from \cite{miyato2018spectral} and \cite{wu2019deep} respectively. We report the Inception Score (IS, higher is better, \citealt{salimans2016improved}) and Fr{\'e}chet Inception Distance (FID, lower is better, \citealt{heusel2017gans}).}
\label{tab:app}
\begin{center}
\begin{tabular}{l|lll}
         & SN-GAN & CS-GAN & LOGAN (NGD) \\
\hline
FID  & $29.3$ & $23.1 \pm 0.5$ & $\mathbf{17.7 \pm 0.4}$ \\
IS   & $7.42 \pm 0.08$ & $7.80 \pm 0.05$ & $\mathbf{8.67 \pm 0.05}$ \\
\end{tabular}
\end{center}
\label{tab:dcgan-scores}
\end{table}

\subsection{Experiments with BigGAN on ImageNet}
\label{sec:biggan}

To illustrate the scalability of our algorithm, we next focus on large scale models based on BigGAN-deep \citep{brock2018large} trained on $128 \times 128$ size images from the ImageNet dataset \citep{imagenet_cvpr09}.

\subsubsection{Model Configuration}

We used the standard BigGAN-deep architecture with three minor modifications:
1. We increased the size of the latent source from $128$ to $256$, to compensate the randomness of the source lost when optimising $z$. 2. We use the uniform distribution $\mathcal{U}(-1, 1)$ instead of the standard normal distribution $\mathcal{N}(0, 1)$ for $p(z)$ to be consistent with the clipping operation (Algorithm \ref{alg}). 3. We use \texttt{leaky~ReLU} (with the slope of $0.2$ for the negative part) instead of \texttt{ReLU} as the non-linearity for smoother gradient flow for $\frac{\partial \fz}{\partial z}$. 

Consistent with the detailed findings in \cite{brock2018large}, our experiment with this baseline model obtains only slightly better scores compared with those in \cite{brock2018large} (Table~\ref{tab:scores}, see also Figure~\ref{fig:all_truncate} in Appendix~\ref{app:samples_and_truncation}). We computed the FID and IS as in \cite{brock2018large}, and computed IS values from checkpoints with the lowest FIDs. Finally, we computed the means and standard deviations for both measures from 5 models with different random seeds.

To apply latent optimisation with NGD, we use the same large step size of $\alpha = 0.9$ as in SN-GAN (section~\ref{app:dcgan}). However, we found much heavier damping is essential for BigGAN, so we use the damping factor $\beta = 5.0$, and only optimise $50\%$ of $z$'s elements. Consistent with \citet{tanaka2019discriminator}, we found a much larger weight of $300.0$ for the regulariser $R_z$ (eq.~\ref{eq:z_reg}) works best, since deeper models generally have larger Lipschitz constants. All other hyper-parameters, including learning rates and a large batch size of 2048, remain the same as in BigGAN-deep. We call this model LOGAN (NGD).

\subsubsection{Basic Results}
\label{sec:biggan-basic}

Employing the same architecture and number of parameters as the BigGAN-deep baseline, LOGAN (NGD) achieved better FID and IS (Table~\ref{tab:scores}). As observed by \citet{brock2018large}, BigGAN training eventually collapsed in every experiment. Training with LOGAN also collapsed, perhaps due to higher-order dynamics beyond the scope we have analysed, but it took significantly longer (600k steps versus 300k steps with BigGAN-deep). 

During training, LOGAN was about $3$ times slower per step compared with BigGAN-deep because of the additional forward and backward passes. In contrast to experiments with smaller models (section~\ref{app:dcgan}), we found that optimising $z$ during evaluation did not improve sample scores (even up to 10 steps), so we do not optimise $z$ for evaluation. Therefore, LOGAN has the same evaluation cost as original BigGAN-deep. To help understand this behaviour, we plot the change from $\Dz$ during training in Figure~\ref{fig:train} \textbf{a}. Although the movement in Euclidean space $\norm{\Dz}$ grew until training collapsed, the movement in $D$'s output space, measured as $\norm{f(z + \Dz) - f(z)}$, remained unchanged (see Appendix~\ref{app:comp-distance} for details). As shown in our analysis, optimising $z$ improves the training dynamics, so LOGANs work well after training without requiring latent optimisation. 
We reckon that smaller models might not be ``over-parametrised'' enough to fully amortise the computation from optimising $z$, which can then further exploit the architecture in evaluation time.
Appendix~\ref{app:analyse-logan} further illustrates these different behaviours. We aim to further investigate this difference in future studies.

Given the criticism of FID and IS as heuristics metrics for sample distributions, we further measure how these samples directly contribute to downstream classification task via the recently proposed Classification Accuracy Score (CAS, \citealt{ravuri2019classification}). Unlike FID and IS, this metric favours likelihood-based models, which are more likely cover all modes representing different classes. The CAS from LOGAN nearly halved the gap between the state-of-the-art GANs and VQ-VAE2 \citep{razavi2019generating}. See Appendix~\ref{app:cas} for more details.

\subsubsection{Ablation Studies}
\label{sec:exp-ablation}

\begin{figure}[htb]
    \centering
    \begin{tabular}{cc}
    \includegraphics[width=0.4\linewidth]{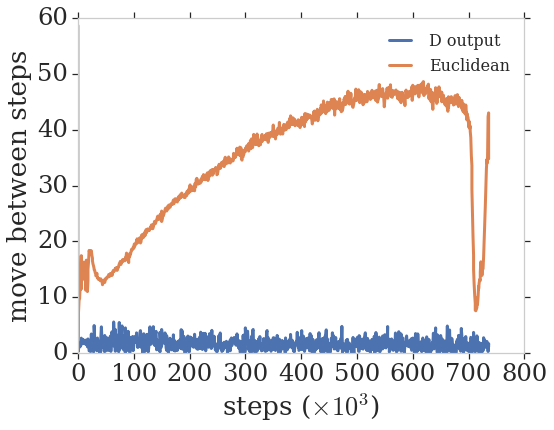} &
    \includegraphics[width=0.4\linewidth]{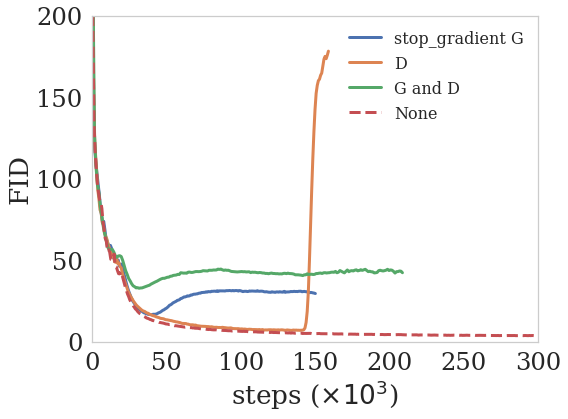} \\
    (\textbf{a}) &
    (\textbf{b})
    \end{tabular}
    \caption{(\textbf{a}) The change from $\Dz$ across training, in $D$'s output space and $z$'s Euclidean space. The distances are normalised by their standard derivations computed from a moving window of size $20$ ($1007$ data points in total). (\textbf{b}) Training curves from models with different ``\texttt{stop\_gradient}'' operations. For reference, the training curve from an unablated model is plotted as the dashed line. All instances with \texttt{stop\_gradient} collapsed (FID went up) early in training.}
    \label{fig:train}
\end{figure}

We verify our theoretical analysis in section~\ref{sec:analysis} by examining key components of Algorithm~\ref{alg} via ablation studies. First, we experiment with using basic GD to optimising $z$, as in \cite{wu2019deep}, and call this model LOGAN (GD). A smaller step size of $\alpha = 0.0001$ was required; larger values were unstable and led to premature collapse of training. As shown in Table~\ref{tab:scores}, the scores from LOGAN (GD) were worse than LOGAN (NGD) and similar to the baseline model.

We then evaluate the effects of removing those terms depending on $\frac{\partial \fz}{\partial z}$ in eq.~\ref{eq:lo-g}, which are not in the ordinary gradient (eq.~\ref{eq:sg}). Since we computed these terms by back-propagating through the latent optimisation procedure, we removed them by selectively blocking back-propagation with ``\texttt{stop\_gradient}'' operations (e.g., in TensorFlow, \citealt{abadi2016tensorflow}). Figure~\ref{fig:train} \textbf{b} shows the change of FIDs for the three models corresponding to removing $\left(\frac{\partial \Delta z}{\partial \theta_G} \right)^T \, \frac{\partial f(z')}{\partial z'}$, removing $\left(\frac{\partial \Delta z}{\partial \theta_D} \right)^T \, \frac{\partial f(z')}{\partial z'}$ and removing both terms.
As predicted by our analysis (section \ref{sec:analysis}), both terms help stabilise training; training diverged early for all three ablations.

\subsubsection{Truncation and Samples}

Truncation is a technique introduced by \cite{brock2018large} to illustrate the trade-off between the FID and IS in a trained model. For a model trained with $z \sim p(z)$ from a source distribution symmetric around $0$, such as the standard normal distribution $\mathcal{N}(0, 1)$ and the uniform distribution $\mathcal{U}(-1, 1)$, down-scaling (truncating) the source $\bar{z} = s \cdot z$ with $0 \leq s < 1$ gives samples with higher visual quality but reduced diversity. We see this quantified in higher IS scores and lower FID when evaluating samples from truncated distributions.

Figure~\ref{fig:model-truncation} \textbf{b} plots the truncation curves for the baseline BigGAN-deep model, LOGAN (GD) and LOGAN (NGD), obtained by varying the truncation (value of $s$) from $1.0$ (no truncation, upper-left ends of the curves) to $0.02$ (extreme truncation, bottom-right ends). Each curve shows the trade-off between FID and IS for an individual model; curves towards the upper-right corner indicate better overall sample quality. The relative positions of curves in figure~\ref{fig:model-truncation} (\textbf{b}) shows LOGAN (NGD) has the best sample quality. Interestingly, although LOGAN (GD) and the baseline model have similar scores without truncation (upper-left ends of the curves, see also Table~\ref{tab:scores}), LOGAN (GD) was better behaved with increasing truncation, suggesting LOGAN (GD) still converged to a better equilibrium. For further reference, we plot truncation curves from additional baseline models in Figure~\ref{fig:all_truncate} (Appendix~\ref{app:samples_and_truncation}).

Figure~\ref{fig:samples-high-IS} and Figure~\ref{fig:samples-low-FIDs} show samples from selected points along the truncation curves. 
In the high IS regime, C and D on the truncation curves both have similarly high IS of near 260. Samples from batches with such high IS have almost photo-realistic image quality. Figure~\ref{fig:samples-high-IS} shows that while the baseline model produced nearly uniform samples, LOGAN (NGD) could still generate highly diverse samples.
On the other hand, A and B from Figure~\ref{fig:model-truncation} \textbf{b} have similarly low FID of near 5, indicating high sample diversity. Samples in Figure~\ref{fig:samples-low-FIDs} \textbf{b} show higher quality compared with those in \textbf{a} (e.g., the interfaces between the elephants and ground, the contours around the pandas).

\section{Conclusion}

In this work, we present LOGAN, which significantly improves the state of the art in large scale GAN training for image generation by optimising the latent source $z$.
Our results illustrate improvements in quantitative evaluation and samples with higher quality and diversity. 
Moreover, our analysis suggests that LOGAN fundamentally improves adversarial training dynamics.
LOGAN is related to the energy-based formulation of a GAN's discriminator \citep{dai2017calibrating,kumar2019maximum,du2019implicit}, when latent optimisation is viewed as descending the energy function defined by the discriminator.
From this view, sampling from the distribution implicitly defined by this energy function, via, e.g., Langevin sampling \citep{welling2011bayesian}, may bring further benefits.
Another class of approaches regularises the entropy of the generator outputs to reduce mode collapse~\citep{mine,presgan}.
Such techniques could be combined with LOGAN to further improve coverage of the underlying data distribution.
Moreover, we expect our method to be useful in other tasks that involve adversarial training, including representation learning and inference \citep{donahue2016adversarial,ali,bigbigan}, text generation \citep{zhang2019self}, style learning \citep{zhu2017unpaired,karras2019style}, audio generation \citep{donahue2018adversarial} and video generation \citep{vondrick2016generating,clark2019efficient}.

\subsubsection*{Acknowledgments}
We thank Mihaela Rosca, Suman Ravuri and James Martens for comments on the draft and insightful discussions.

\bibliography{iclr2020_conference}
\bibliographystyle{iclr2020_conference}


\appendix

\section{Detailed Analysis of Latent Optimisation}
\label{app:analysis}
In this section we present complementary of LOGAN. In particular, we show how the algorithm brings together ideas from symplectic gradient adjustment and stochastic approximation with two time scales. 

\subsection{Approximate Symplectic Gradient Adjustment}
\label{app:approx-sga}

To analyse LOGAN as a differentiable game we treat the latent step $\Delta z$ as adding a \emph{third player} to the original game played by the discriminator and generator. The third player's parameter, $\Dz$, is optimised online for each $z \sim p(z)$. Together the three players (latent player, discriminator, and generator) have losses \emph{averaged over a batch of samples}:
\begin{equation}
    L = [\eta \, L_G, L_D, L_G]^T
\end{equation}
where $\eta = \frac{1}{N}$ ($N$ is the batch size) reflects the fact that each $\Dz$ is only optimised for a single sample $z$, so its contribution to the total loss across a batch is small compared with $\pD$ and $\pG$ which are directly optimised for batch losses. This choice of $\eta$ is essential for the following derivation, and has important practical implication. It means that the per-sample loss $L_G(z')$, instead of the loss summed over a batch $\sum_{n=1}^N L_G(z'_n)$, should be the only loss function guiding latent optimisation. Therefore, when using natural gradient descent (Section~4), the Fisher information matrix should only be computed using the current sample $z$.

The resulting simultaneous gradient is
\begin{equation}
\begin{split}
    g &= \begin{bmatrix}
    \eta \, \frac{\partial L_G(z')}{\partial \Dz},
    \frac{\partial L_D(z')}{\partial \theta_D}, 
    \frac{\partial L_G(z')}{\partial \theta_G} 
    \end{bmatrix}^T \\ 
    &= \begin{bmatrix}
    - \eta \, \frac{\partial f(z')}{\partial \Dz},
    \frac{\partial f(z')}{\partial \theta_D}, 
    - \frac{\partial f(z')}{\partial \theta_G}
    \end{bmatrix}^T \label{eq:sg-lo}
\end{split}
\end{equation}
Following \cite{balduzzi2018mechanics}, we can write the Hessian of the game as:
\begin{equation}
H =
    \begin{bmatrix}
    -\eta \, \frac{\partial^2 \fzp}{\partial \Dz^2} & 
    -\eta \, \frac{\partial^2 \fzp}{\partial \Dz \partial \pD} &
    -\eta \, \frac{\partial^2 \fzp}{\partial \Dz \partial \theta_G}
    \\
    \frac{\partial^2 \fzp}{\partial \pD \partial \Dz} &
    \frac{\partial^2 \fzp}{\partial \pD^2} &
    \frac{\partial^2 \fzp}{\partial \theta_D \, \partial \theta_G}
    \\
    -\frac{\partial^2 \fzp}{\partial \theta_G  \partial \Dz} &
    -\frac{\partial^2 \fzp}{\partial \theta_G \partial \pD} &
    -\frac{\partial^2 \fzp}{\partial \pG^2}
    \end{bmatrix}
\end{equation}

The presence of a non-zero anti-symmetric component in the Hessian
\begin{equation}
\begin{split}
A &= \frac{1}{2}(H - H^T)\\ 
&= \begin{bmatrix}
    0 & 
    -\frac{1 + \eta}{2}\,\frac{\partial^2 \fzp}{\partial \Dz \partial \pD} &
    \frac{1 - \eta}{2} \, \frac{\partial^2 \fzp}{\partial \Delta z \partial \pG}
    \\
    \frac{1 + \eta}{2}\,\frac{\partial^2 \fzp}{\partial \pD \partial \Dz} &
    0 &
    \frac{\partial^2 \fzp}{\partial \theta_D \, \partial \theta_G}
    \\
    -\frac{1 - \eta}{2} \, \frac{\partial^2 \fzp}{\partial \pG \partial \Dz} &
    -\frac{\partial^2 \fzp}{\partial \theta_G \partial \pD} &
    0
\end{bmatrix}
\end{split}
\end{equation}
implies the dynamics have a rotational component which can cause cycling or slow down convergence. 
Since $\eta \ll 1$ for typical batch sizes (e.g., $\frac{1}{64}$ for DCGAN and $\frac{1}{2048}$ for BigGAN-deep), we abbreviate $\gamma = \frac{1 + \eta}{2} \approx \frac{1 - \eta}{2}$ to simplify notations. 

Symplectic gradient adjustment (SGA) counteracts the rotational force by adding an adjustment term to the gradient to obtain $g^* \leftarrow g + \lambda \, A^T g $, which for the discriminator and generator has the form:
\begin{align}
    g^*_D &=  \frac{\partial \fzp}{\partial \theta_D} 
    + \lambda \gamma \, \left( \frac{\partial^2 \fzp}{\partial \Dz \partial \pD}\right)^T \, \frac{\partial \fzp}{\partial \Dz} + \lambda \, \left(\frac{\partial^2 f(z')}{\partial \theta_G \, \partial \theta_D} \right)^T \, \frac{\partial f(z')}{\partial \theta_G} \label{eq:sga-d} \\
    g^*_G &=  -\frac{\partial f(z')}{\partial \theta_G}
    - \lambda \gamma \, \left( \frac{\partial^2 \fzp}{\partial \Dz \partial \pG} \right)^T\, \frac{\partial \fzp}{\partial \Dz} + \lambda \, \left(\frac{\partial f(z')}{\partial \theta_D \, \partial \theta_G} \right)^T \, \frac{\partial f(z')}{\partial \theta_D}\label{eq:sga-g}
\end{align}
The gradient with respect to $\Dz$ is ignored since the convergence of training only depends on $\pD$ and $\pG$. 

If we drop the last terms in eq.\ref{eq:sga-d} and \ref{eq:sga-g}, which are expensive to compute for large models with high-dimensional $\pD$ and $\pG$, and use $\frac{\partial \fzp}{\partial \Dz} = \frac{\partial  \fzp}{\partial z'}$, the adjusted updates can be rewritten as
\begin{align}
    g^*_D &\approx  \frac{\partial f(z')}{\partial \theta_D} 
    + \lambda \gamma \, \left( \frac{\partial^2 \fzp}{\partial z' \partial \pD}\right)^T \, \frac{\partial \fzp}{\partial z'}
    \label{eq:asga-d} \\
    g^*_G &\approx  -\frac{\partial f(z')}{\partial \theta_G}
    - \lambda \gamma \, \left( \frac{\partial^2 \fzp}{\partial z' \partial \pG} \right)^T\, \frac{\partial \fzp}{\partial z'} 
    \label{eq:asga-g}
\end{align}
Because of the third player, there are still the terms depend on $\frac{\partial \fzp}{\partial z'}$ to adjust the gradients. Efficiently computing $\frac{\partial^2 \fzp}{\partial z' \partial \pD}$ and $\frac{\partial^2 \fzp}{\partial z' \partial \pD}$ is non-trivial (e.g., \citealt{pearlmutter1994fast}). 
However, if we introduce the local approximation
\begin{equation}
\frac{\partial^2 \fzp}{\partial z' \partial \pD} \approx \frac{\partial^2 \fz}{\partial z \partial \pD} \qquad \frac{\partial^2 \fzp}{\partial z' \partial \pD} \approx \frac{\partial^2 \fz}{\partial z \partial \pD}
\label{eq:sga-z-approx}
\end{equation}
then the adjusted gradient becomes identical to eq.~6 from latent optimisation.

In other words, automatic differentiation by commonly used machine learning packages can compute the adjusted gradient for $\pD$ and $\pG$ when back-propagating through the latent optimisation process.
Despite the approximation involved in this analysis, both our experiments in section~5 and the results from \cite{wu2019deep} verified that latent optimisation can significantly improve GAN training. 

\subsection{Stochastic Approximation with Two Time Scales}
\label{sec:a-stoch-approx}

\begin{figure}[ht]
    \centering
    \begin{tabular}{c}
    \includegraphics[width=0.5\linewidth]{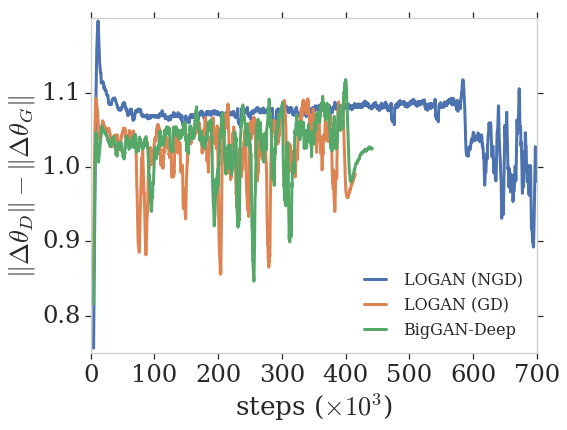}
    \end{tabular}
    \caption{The update speed of the discriminator relative to the generator shown as the difference $\norm{\Delta \theta_D} - \norm{\Delta \theta_G}$ after each update step. Lines are smoothed with a moving average using window size 20 (in total, there are 3007, 1659 and 1768 data points for each curve). All curves oscillated strongly after training collapsed.}
    \label{fig:grad-scale-update}
\end{figure}

This section shows that latent optimisation accelerates the speed of updating $D$ relative to the speed of updating $G$, facilitating convergence according to~\citet{heusel2017gans} (see also Figure~\ref{fig:grad-scale-update} \textbf{b}). Intuitively, the generator  requires less updating compared with $D$ to achieve the same reduction of loss because latent optimisation ``helps'' $G$.

\cite{heusel2017gans} used the theory of stochastic approximation to analyse GAN training. Viewing the training process as stochastic approximation with two time scales \citep{borkar1997stochastic,konda1999actor}, they suggest that the update of $D$ should be fast enough compared with that of $G$. Under mild assumptions, \cite{heusel2017gans} proved that such two time-scale update converges to local Nash equilibrium.
Their analysis follows the idea of $(\tau, \delta)$ perturbation \citep{hirsch1989convergent}, where the slow updates ($G$) are interpreted as a small perturbation over the ODE describing the fast update ($D$). Importantly, the size of perturbation $\delta$ is measured by the magnitude of parameter change, which is affected by both the learning rate and gradients.

Here we show, in accordance with \cite{heusel2017gans}, that LOGAN accelerates discriminator updates and slows down generator updates, thus helping the convergence of discriminator.
We start from analysing the change of $\pG$. We assume that, without LO, it takes $\Delta \theta_G = \theta'_G - \theta_{G}$ to make a small constant amount of reduction in loss $L_G$: 
\begin{equation}
    \rho =  - f(z; \theta_D, \theta_G + \Delta \pG) + f(z; \theta_D, \theta_G)
    \label{eq:rho_1}
\end{equation}
Now using the optimised $z' = z + \Delta z$, we assess the change $\delta \theta_G $ required to achieve the same amount of reduction:
\begin{equation}
\rho = - f(z + \Delta z; \theta_D, \theta_G + \delta \theta_G) + f(z; \theta_D, \theta_G)
    \label{eq:rho_2}
\end{equation}
Intuitively, when $z$ ``helps'' $\theta_G$ to achieve the same goal of increasing $\ff$ by $\rho$, the responsible of $\theta_G$ becomes smaller, so it does not need to change as much as $\Delta \theta_G$, thus $\norm{\delta \theta_G} < \norm{\Delta \theta_G}$.

Formally,  $\ff$ and $f(z + \Delta; \theta_D, \theta_G + \delta \theta_G)$ have the following Taylor expansions around $z$ and $\theta_G$:
\begin{align}
    f(z; \theta_d, \theta_G + \delta \theta_G) =& \ff 
    + \left( \frac{\partial \ff}{\partial \pG} \right)^T \, \Delta \pG \label{eq:f-o-taylor} + \epsilon(\Delta \pG) \\
    f(z + \Delta z; \theta_d, \theta_G + \delta \theta_G)  =& \ff  + \left( \frac{\partial \ff}{\partial z} \right)^T \Delta z \\
    &\, + \left( \frac{\partial \ffe{+\Delta z}{}{}}{\partial \theta_G} \right)^T \, \delta \theta_G + \epsilon(\Delta z, \delta \pG)
\label{eq:f-taylor}
\end{align}
Where $\epsilon(\cdot)$'s are higher order terms of the increments.
Using the assumption of eq.~\ref{eq:rho_1} and \ref{eq:rho_2}, we can combine eq.~\ref{eq:f-o-taylor} and \ref{eq:f-taylor}:
\begin{equation}
    \left(\frac{\partial \ff}{\partial \pG}\right)^T \, \Delta \pG = \left( \frac{\partial \ff}{\partial z} \right)^T \, \Delta z + \left( \frac{\partial \ffe{+ \Delta z}{}}{\partial \theta_G} \right)^T \, \delta \theta_G + \epsilon
    \label{eq:pG-comb}
\end{equation}
where $\epsilon = \epsilon(\Delta z, \delta \pG) - \epsilon(\Delta \pG)$. Since $\Delta z \propto \frac{\partial \ff}{\partial z}$ in gradient descent (eq.~3), 
\begin{equation}
\frac{\partial \ff}{\partial z} \, \Delta z > 0
\label{eq:z-step}
\end{equation}
Therefore, we have the inequality
\begin{equation}
    \left( \frac{\partial \ff}{\partial \pG} \right)^T \, \Delta \pG < \left(\frac{\partial \ffe{+ \Delta z}{}}{\partial \theta_G}\right)^T \, \delta \theta_G + \epsilon
    \label{eq:ieq_1}
\end{equation}
If we further assume $\Delta \pG$ and  $\delta \pG$ are obtained from stochastic gradient descent with identical learning rate,
\begin{equation}
\Delta \pG = \alpha \, \frac{\partial \ff}{\partial \pG} \qquad
\delta \pG = \alpha \, \frac{\partial \ff}{\partial \pG}
\label{eq:sgd}
\end{equation}
substituting eq.~\ref{eq:sgd} into eq.~\ref{eq:ieq_1} gives
\begin{equation}
    \norm{\Delta \pG} < \norm{\delta \pG} + \epsilon
\end{equation}
The same analysis applies to the discriminator. The similar intuition is that it takes the discriminator additional effort to compensate the exploitation from the optimised $z'$. We then obtain
\begin{equation}
    \left(\frac{\partial \ff}{\partial \pD}\right)^T \, \Delta \pD = \left( \frac{\partial \ff}{\partial z} \right)^T \, \Delta z + \left( \frac{\partial \ffe{+ \Delta z}{}}{\partial \theta_D} \right)^T \, \delta \theta_D + \epsilon
\end{equation}
However, since the adversarial loss $L_D = -L_G$, we have
$\Delta \pD = -\alpha \, \frac{\partial \ff}{\partial \pD}$ and
$\delta \pD = -\alpha \, \frac{\partial \ff}{\partial \pD}$
taking the opposite signs of eq.\ref{eq:sgd}.  
For sufficiently small $\Delta z$, $\Delta \pG$ and $\delta \theta_G$, $\epsilon$ is close to zero, 
so $\norm{\Delta \pD} < \norm{\delta \pD}$ under our assumptions of small $\Delta z$, $\Delta \pG$ and $\delta \theta_G$.

Importantly, the bigger the product $\frac{\partial \fz}{\partial z} \, \Dz$ is, the more robust the inequality is to the error from $\epsilon$. 
Moreover, bigger steps increase the speed gap between updating D and G, further facilitating convergence (in accordance with~\cite{heusel2017gans}). Overall, our analysis suggests:
\begin{enumerate}
    \item More than one gradient descent step may not be helpful, since $\Dz$ from multiple GD steps may deviate from the direction of $\frac{\partial \fz}{\partial z}$.
    \item A large step of $\Dz$ is helpful in facilitating convergence by widening the gap between D and G updates \citep{heusel2017gans}.
    \item However, the step of $\Dz$ cannot be too large. In addition to the linear approximation we used throughout our analysis, the approximate SGA breaks down when eq.\ref{eq:sga-z-approx} is strongly violated when ``overshoot'' brings the gradients at $\frac{\partial \fzp}{\partial z'}$ to the opposite sign of $\frac{\partial \fz}{\partial z}$.
\end{enumerate}

\section{Additional Analysis of Latent Optimisation}
\label{app:analyse-logan}

Here we analyse the relationship between the number of latent optimisation steps during evaluation and the final FIDs and inception scores. As in the main paper, we train the SN-GAN model with only 1 latent optimisation step, but test them with $\{ 0, 1, 5, 10, 20, 30\}$ steps during evaluation. For lower variance in computing the scores, we use $10,000$ samples for evaluation (as oppose to $5000$ samples used in the main paper for direct comparison with other baselines). The inception scores are taken from checkpoints with the best (lowest) FIDs, and the error bars indicate standard deviations obtained from 3 different random seeds.

\begin{figure}[ht]
    \centering
    \begin{tabular}{cc}
    \includegraphics[width=0.45\linewidth]{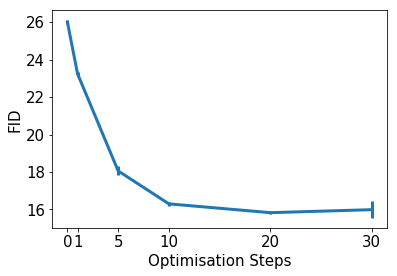} &
    \includegraphics[width=0.45\linewidth]{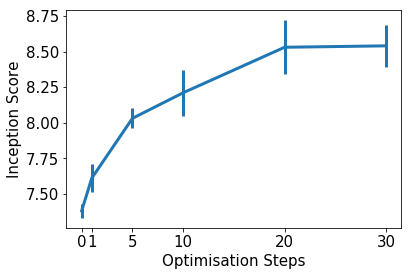} \\
    (\textbf{a}) &
    (\textbf{b}) 
    \end{tabular}
    \caption{FIDs (\textbf{a}) and Inception Scores (\textbf{b}) obtained with different latent optimisation steps \emph{at evaluation}.}
    \label{fig:lo-scores}
\end{figure}

Figure~\ref{fig:lo-scores} shows that the scores can be substantially improved with extra optimisation steps in evaluation. Although only 1 step was used in training, up to around 20 steps at evaluation could still improve sample quality. Beyond that, the return from extra computation became diminishing.

\begin{figure}[ht]
    \centering
    \begin{tabular}{ccc}
    \includegraphics[width=0.3\linewidth]{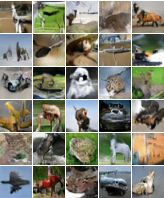} &
    \includegraphics[width=0.3\linewidth]{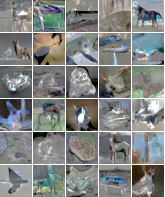} &
    \includegraphics[width=0.3\linewidth]{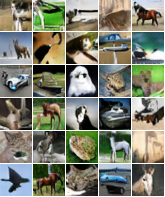} \\
    \textbf{before} &
    \textbf{difference} &
    \textbf{after} 
    \end{tabular}
    \caption{CIFAR samples from SN-GAN before, after latent optimisation, and the differences between them.}
    \label{fig:dcgan-lo-diff}
\end{figure}

\begin{figure}[ht]
    \centering
    \begin{tabular}{ccc}
    \includegraphics[width=0.3\linewidth]{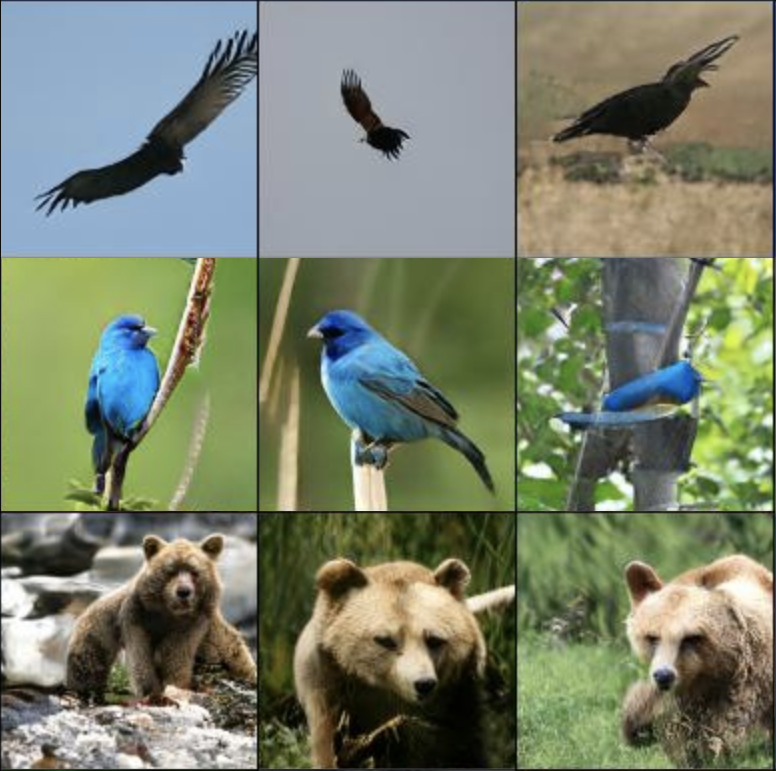} &
    \includegraphics[width=0.3\linewidth]{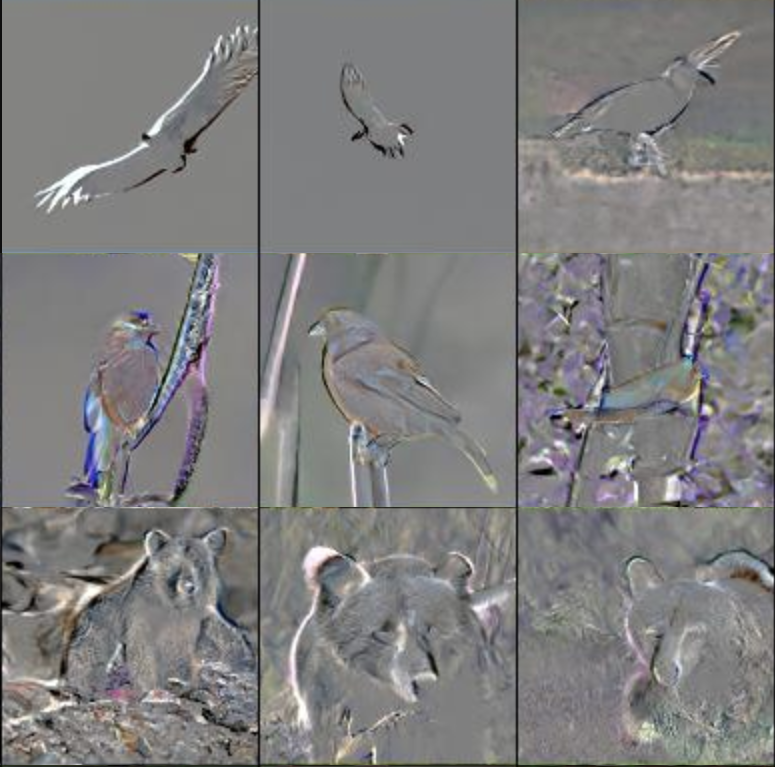} &
    \includegraphics[width=0.3\linewidth]{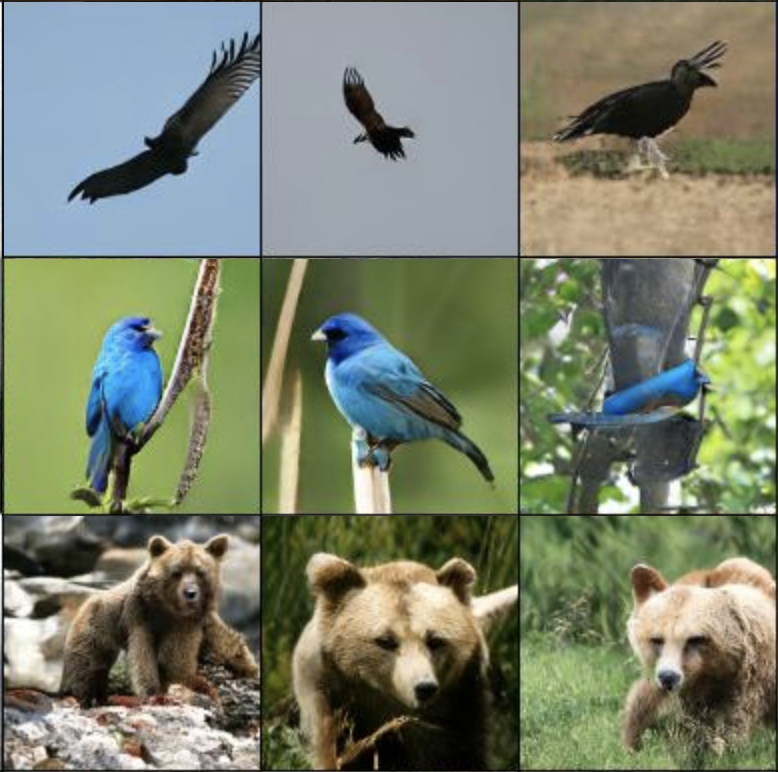} \\
    \textbf{before} &
    \textbf{difference} &
    \textbf{after} 
    \end{tabular}
    \caption{ImageNet samples from BigGAN-deep before, after latent optimisation, and the differences between them.}
    \label{fig:biggan-lo-diff}
\end{figure}

We did not observe similar improvement with BigGANs in evaluation. To contrast the difference between them, Figure~\ref{fig:dcgan-lo-diff} and \ref{fig:biggan-lo-diff} illustrate the change of samples made by latent optimisation. In both cases, 10 latent optimisation steps were applied in evaluation, after the models were trained with 1 optimisation step. While the effect of improvement is clear in Figure~\ref{fig:dcgan-lo-diff} (from SN-GAN), the changes in Figure~\ref{fig:biggan-lo-diff} are barely observable from inspecting the samples alone before and after latent optimisation.

\section{Evaluating Classification Accuracy Score}
\label{app:cas}

To compute CAS, a ResNet classifier \citep{he2016deep} is trained on samples from our model instead of the ImageNet dataset. Each data sample is replaced by a sample from the model conditioned on the same class. The trained model is then evaluated on the ImageNet dataset as in a standard classification task. We use the same schedule as in \cite{ravuri2019classification} for training the ResNet classifier, but stopped earlier at about 10k steps, where the classification accuracy peaked. See \cite{ravuri2019classification} for the motivation of CAS and more details of the training and evaluation procedure. We report both top-5 and top-1 classification accuracy in Table~\ref{tab:cas}. Although higher resolution generally brings better CAS, we use the $128 \times 128$ model as in the main paper due to limited computational resource. Despite this, the CAS from LOGAN nearly halved the gaps between BigGAN-deep and VQ-VAE2 at a higher resolution of $256 \times 256$.

\begin{table}[htb]
\caption{CAS for different models. Except LOGAN, numbers from all other models are reproduced from \cite{razavi2019generating}.}
\label{tab:cas}
\begin{center}
\begin{tabular}{l|c|c}
         & Top-5 Accuracy & Top-1 Accuracy \\
\hline
BigGAN-deep ($128 \times 128$)  & $64.44\%$ & $40.64\%$ \\
BigGAN-deep ($256 \times 256$) & $65.92\%$ & $42.65\%$\\
LOGAN ($128 \times 128$)  & $71.97\%$ & $47.86\%$ \\
VQ-VAE2 ($256 \times 256$)  & $77.59\%$ & $54.83\%$ \\
Real Data ($256 \times 256$)  & $88.79\%$ & $68.82\%$ \\
\end{tabular}
\end{center}
\end{table}

\section{Poisson Likelihood from Hinge loss}
\label{app:poisson-ll}

Here we provide a probabilistic interpretation of the hinge loss for the generator, which leads naturally to the scenario of a family of discriminators. Although this interpretation is not necessary for our current algorithm, it may provides useful guidance for incorporating multiple discriminators.

We introduce the label $t=1$ for real data and $t=0$ fake samples.
This section shows that the generator hinge loss
\begin{equation}
    L_G = - D \left( G(z) \right)
\end{equation}
can be interpreted as a negative log-likelihood function:
\begin{equation}
    L_G = - \ln p(t = 1; D, G(z))
\end{equation}
Here $p(t = 1; z, D, G)$ is the probability that the  generated image $G(z)$ can fool the discriminator $D$.

The original GAN's discriminator can be interpreted as outputting a Bernoulli distribution $p(t; \beta_G) = \beta_G^t \cdot (1- \beta_G)^{1-t}$. In this case, if we parameterise $\beta_G = D\left( G(z)\right)$, the generator loss is the negative log-likelihood
\begin{equation}
\begin{split}
    -\ln P\big(t=1; D, G(z)\big) &= -\ln p(t=1; \beta_G) \\
    &= -\ln \beta_G = - \ln D\left( G(z)\right)
\end{split}
\end{equation}

Bernoulli, however, is not the only valid choice as the discriminator's output distribution. Instead of sampling ``1" or ``0", we assume that there are \emph{many} identical discriminators that can independently vote to reject an input sample as fake. The number of votes $k$ in a given interval can be described by a Poisson distribution with parameter $\lambda$ with the following PMF:
\begin{equation}
    p(k; \lambda) = \frac{\lambda^k e^{-\lambda}}{k!}
\end{equation}

The probability that a generated image can fool \emph{all} the discriminators is the probability of $G(z)$ receiving no vote for rejection
\begin{equation}
    p(k=0; \lambda) = e^{-\lambda}
\end{equation}
Therefore, we have the following negative log-likelihood as the generator loss if we parameterise $\lambda = -D \left( G(z)\right)$:
\begin{equation}
    -\ln p\big(k = 0; D, G(z)\big) = -\ln p(k=0; \lambda) = -D \left( G(z)\right)
\end{equation}
This interpretation has a caveat that when $D \left( G(z)\right) > 0$ the Poisson distribution is not well defined. However, in general the discriminator's hinge loss
\begin{equation}
    L_D = - \min\big(0, -1 + D(x)\big) - \min \big(0, -1 - D(G(z))\big)
\end{equation}
pushes $D \left( G(z)\right) < 0$ via training.

\section{Hyper-parameter Search}
\label{app:hyper}

We first searched the hyper-parameters for the DCGAN (section~5.1) over the following range:
\begin{table}[htb]
\caption{Hyper-parameter grid for the DCGAN. Best values from the grid search are highlighted.}
\begin{center}
\begin{tabular}{l|c}
         & values \\
\hline
$\alpha$  & $\{0.01, 0.1, 0.3, 0.5, 0.7, 0.8, \mathbf{0.9}, 1.0\}$ \\
$\beta$  & $\{0.01, \mathbf{0.1}, 0.3, 0.5, 0.7, 0.9, 1.0\}$ \\
$w_r$  & $\{0.01, \mathbf{0.1}, 0.2, 0.3, 0.5, 0.7, 1.0, 2.0, 3.0\}$ \\
$c$  & $\{30\%, 50\%, 70\%, \mathbf{80\%}, 90\%, 100\% \}$ \\
\end{tabular}
\end{center}
\end{table}

Base on the results from DCGANs, hyper-parameter search on the following grid was performed for the BigGAN-deep (section~5.2):
\begin{table}[htb]
\caption{Hyper-parameter grid for the BigGAN-deep. Best values from the grid search are highlighted.}
\begin{center}
\begin{tabular}{l|c}
         & values \\
\hline
$\alpha$  & $\{0.7, 0.8, \mathbf{0.9}, 1.0\}$ \\
$\beta$  & $\{0.1, 0.5, 1.0, 2.0, 3.0, 4.0, \mathbf{5.0}, 7.0, 9.0\}$ \\
$w_r$  & $\{0.1, 0.2, 0.5, 1.0, 2.0, 5.0, 10.0, 30.0$ \\
 & $100.0, 200.0, \mathbf{300.0}, 400.0, 500.0\}$ \\
$c$  & $\{30\%, \mathbf{50\%}, 70\%, 80\%\}$ \\
\end{tabular}
\end{center}
\end{table}

\section{Details in Computing Distances in Figure 5 \textbf{a}}
\label{app:comp-distance}

For a temporal sequence $x_1, x_2, \dots, x_T$ (e.g. the changes of $z$ or $f(z)$ at each training step in this paper),
to normalise its variance while accounting for the non-stationarity, we process it as follows. We first compute the moving average and standard deviation over a window of size $N$:
\begin{align}
    \mu_t &= \frac{1}{N}\sum_{u=t}^{t+N-1} x_u \\
    \sigma_t &= \sqrt{\frac{1}{N - 1}\sum_{u=t}^{t+N-1} (x_u - \mu_u)^2}
\end{align}
Then normalise the sequence as:
\begin{equation}
    \bar{x}_t = \frac{x_t}{\sigma_t}
\end{equation}
The result in Figure 5\textbf{a} is robust to the choice of window size. Our experiments with $N$ from $10$ to $50$ yielded visually similar plots.

\section{Additional Samples and Results}
\label{app:samples_and_truncation}

Figure~1 and 2 provide additional samples, organised similar to Figure~1 and 2. Figure~3 shows additional truncation curves.

\begin{figure}[htb]
    \centering
    \begin{tabular}{cc}
    \includegraphics[width=0.43\linewidth]{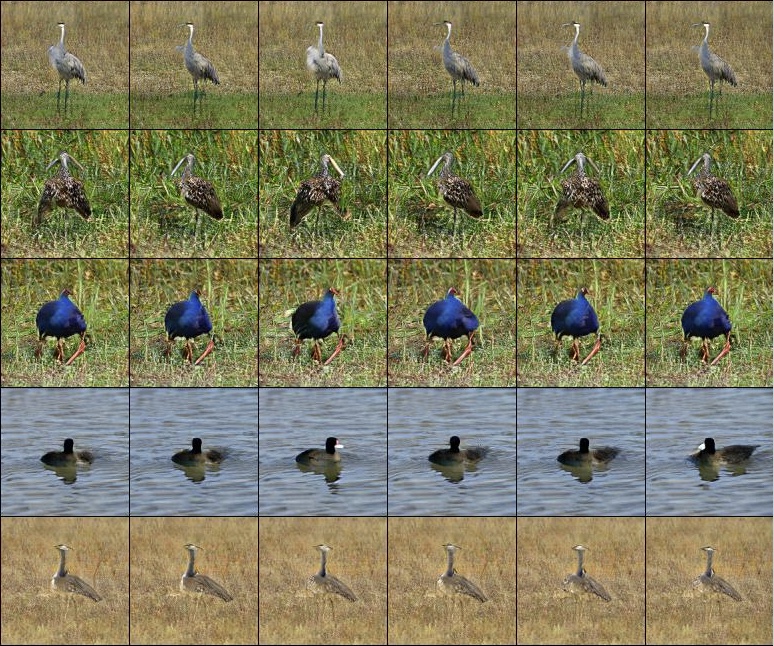}
    &   
    \includegraphics[width=0.43\linewidth]{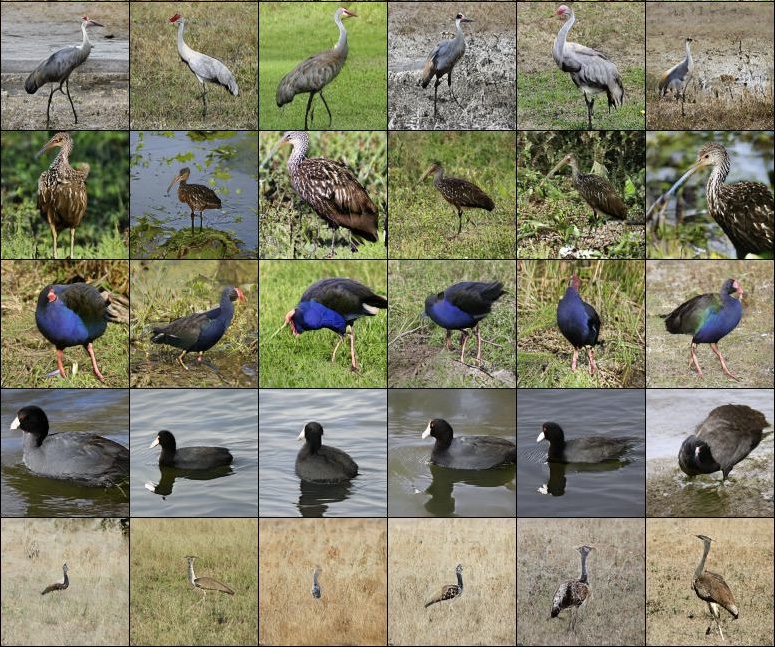} \\
    \includegraphics[width=0.43\linewidth]{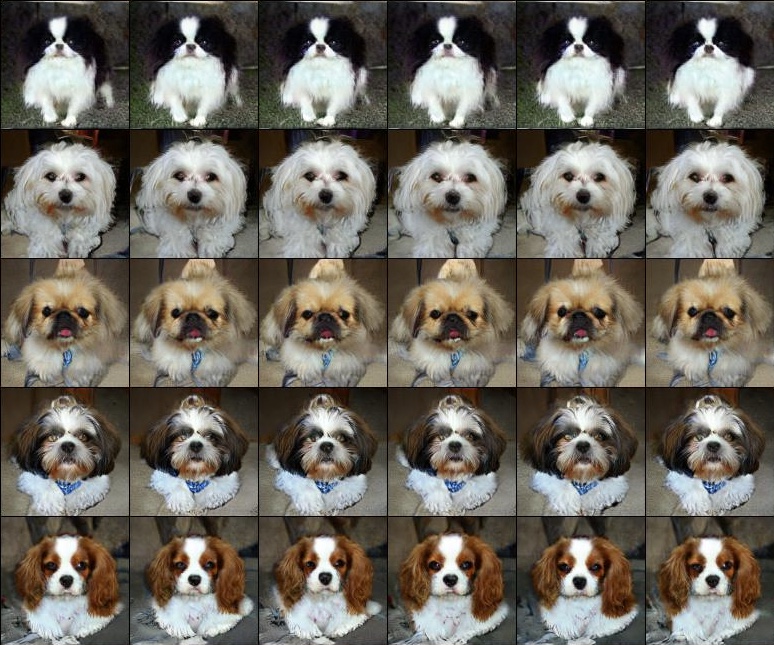}
    &   
    \includegraphics[width=0.43\linewidth]{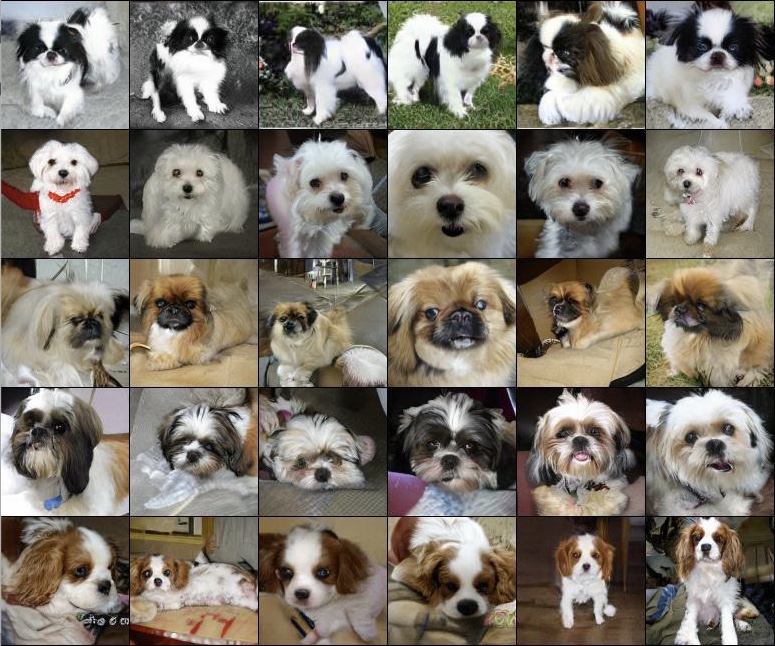} \\
    \includegraphics[width=0.43\linewidth]{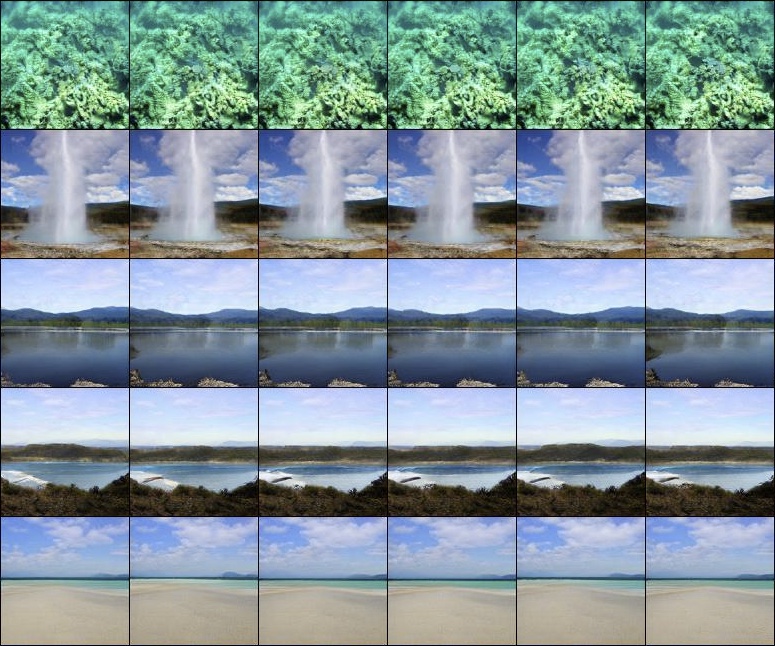}
    &   
    \includegraphics[width=0.43\linewidth]{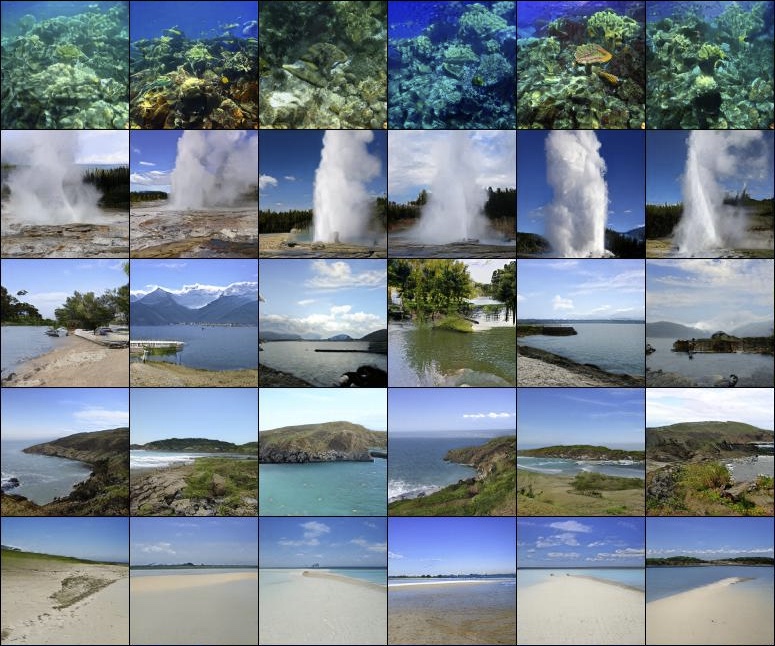} \\
    \includegraphics[width=0.43\linewidth]{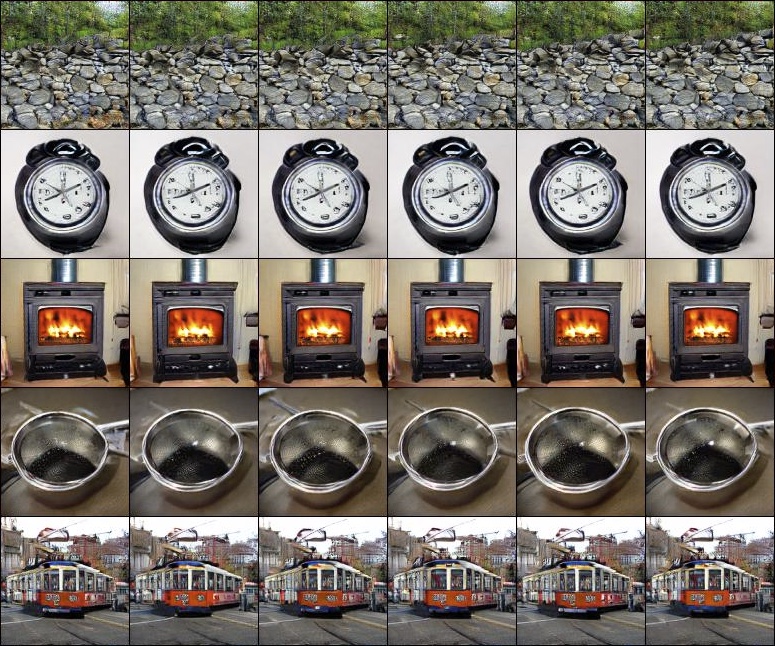}
    &   
    \includegraphics[width=0.43\linewidth]{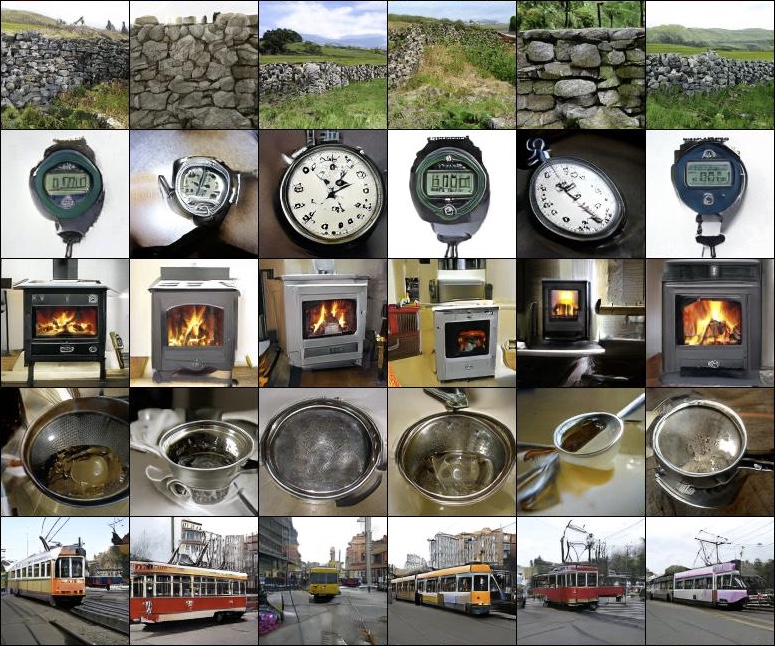} \\
    (\textbf{a}) & (\textbf{b}) \\
    \end{tabular}
    \caption{Samples from BigGAN-deep (\textbf{a}) and LOGAN (\textbf{b}) with similarly high Inception scores. Samples from the two panels were drawn from truncations correspond to points C, D in Figure~3. (FID/IS: (\textbf{a}) 27.97/259.4, (\textbf{b}) 8.19/259.9)}
    \label{fig:app-samples-high-IS}
\end{figure}

\begin{figure}[htb]
    \centering
    \begin{tabular}{cc}
    \includegraphics[width=0.47\linewidth]{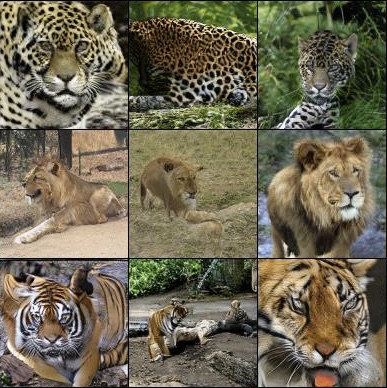}
    &   \includegraphics[width=0.47\linewidth]{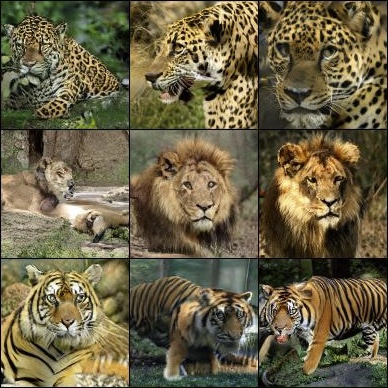} \\
    \includegraphics[width=0.47\linewidth]{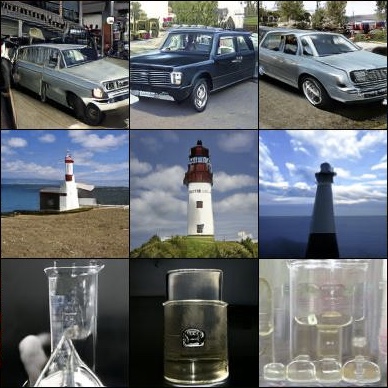}
    &   \includegraphics[width=0.47\linewidth]{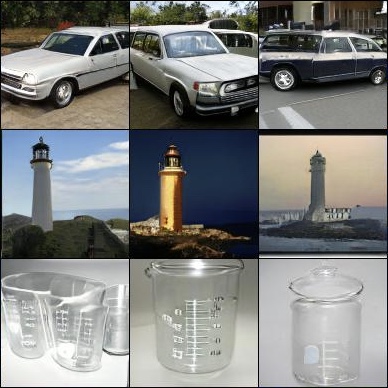} \\
    \includegraphics[width=0.47\linewidth]{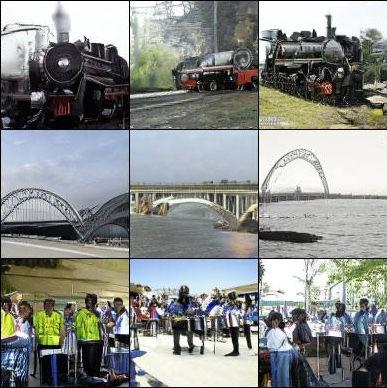}
    &   \includegraphics[width=0.47\linewidth]{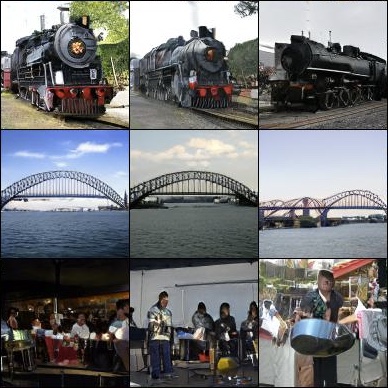} \\
    (\textbf{a}) & (\textbf{b}) 
    \end{tabular}
    \caption{Samples from BigGAN-deep (\textbf{a}) and LOGAN (\textbf{b}) with similarly low FID. Samples from the two panels were draw from truncations correspond to points A, B in figure~3\textbf{b}. (FID/IS: (\textbf{a}) 5.04/126.8, (\textbf{b}) 5.09/217.0)}
    \label{fig:app-samples-low-FIDs}
\end{figure}

\begin{figure}
    \centering
    \includegraphics[width=0.8\linewidth]{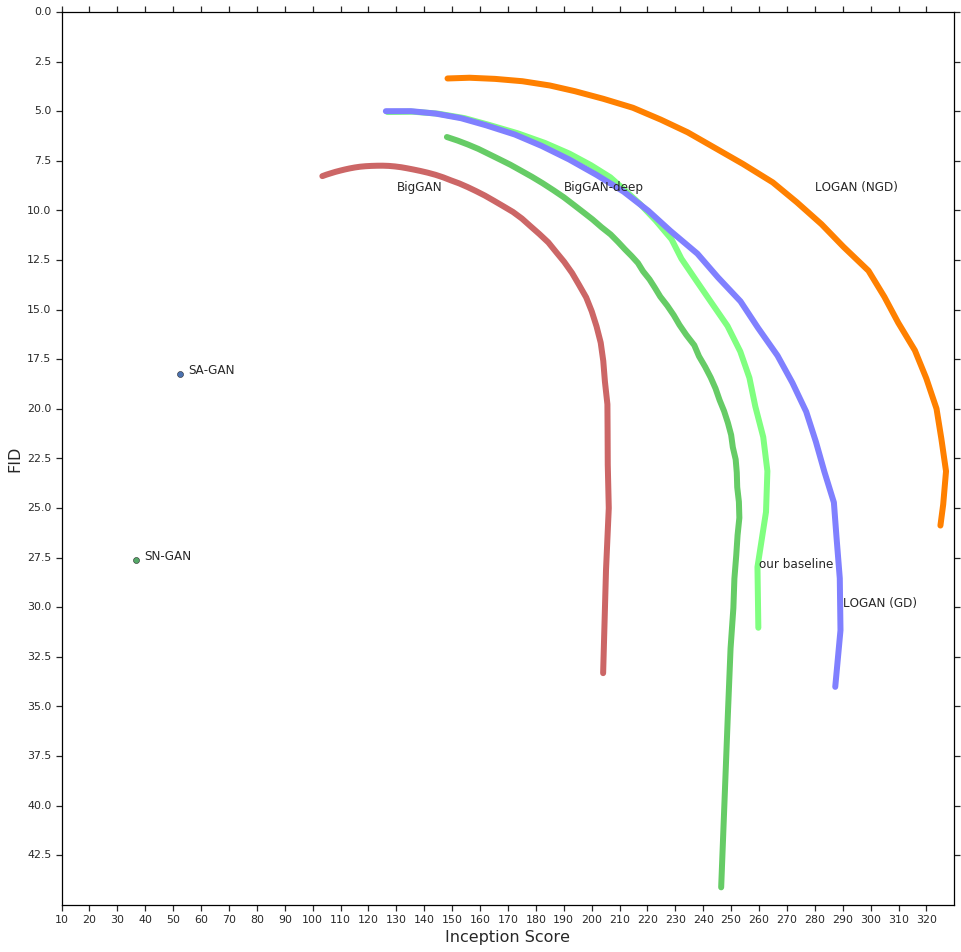}
    \caption{Truncation curves with additional baselines. In addition to the truncation curves reported in Figure~3\textbf{b}, here we also include the Spectral-Normalised GAN \citep{miyato2018spectral}, Self-Attention GAN \citep{zhang2019self}, original BigGAN and BigGAN-deep as presented in \cite{brock2018large}.}
    \label{fig:all_truncate}
\end{figure}


\end{document}